\documentclass[10pt,journal,compsoc]{IEEEtran}
\ifCLASSOPTIONcompsoc
\usepackage[nocompress]{cite}
\else
\usepackage{cite}
\fi
\usepackage{textcomp}
\usepackage[utf8]{inputenc} % allow utf-8 input
\usepackage{stfloats}
\usepackage{url}            % simple URL typesetting
\usepackage{booktabs}       % professional-quality tables
\usepackage{amsfonts}       % blackboard math symbols
\usepackage{nicefrac}       % compact symbols for 1/2, etc.
\usepackage{microtype}      % microtypography
\usepackage{graphicx}
\usepackage{enumitem}
\usepackage{amssymb}
\usepackage{amsmath}
\usepackage{verbatim}
\usepackage[table]{xcolor}
\usepackage{subfig}
\usepackage{multirow}
\usepackage{ragged2e}
\usepackage{overpic}
\usepackage{color}
\usepackage{colortbl} % 用于给表格的行列上色
\definecolor{tabletitle}{gray}{.8}
\definecolor{ours}{gray}{.95}
\definecolor{ggray}{RGB}{127,127,127}
\definecolor{reda}{RGB}{202,0,0}
\definecolor{redb}{RGB}{217,148,143}
\definecolor{myyellow}{RGB}{190,144,0}
\definecolor{mygreen}{RGB}{0,136,51}
\definecolor{myblue}{RGB}{0,102,204}
\newcolumntype{B}{!{\vrule width 1pt}}
\usepackage{pifont}
\usepackage[colorlinks,hyperindex,breaklinks]{hyperref}
\newcommand{\myparagraph}[1]{\vspace{0.1em}\noindent\textbf{#1}}
\newcommand{\ie}{\textit{i}.\textit{e}.}
\newcommand{\eg}{\textit{e}.\textit{g}.}
\begin{document}
% --------------------------------------------
% \title{Complementary Triple-Decoder Network for Robust RGB-T Salient Object Detection and Beyond}
\title{Divide-and-Conquer: Confluent Triple-Flow Network for RGB-T Salient Object Detection}
% --------------------------------------------
\author{Hao~Tang, Zechao~Li, Dong~Zhang, Shengfeng~He, Jinhui~Tang%,~\IEEEmembership{Senior Member,~IEEE}
% --------------------------------------------	
\IEEEcompsocitemizethanks{
\IEEEcompsocthanksitem{H. Tang, Z. Li, and J. Tang are with the School of Computer Science and Engineering, Nanjing University of Science and Technology, No 200 Xiaolingwei, Nanjing 210094, China. E-mail: \{tanghao0918, zechao.li, jinhuitang\}@njust.edu.cn. }
\IEEEcompsocthanksitem {D. Zhang is with the Department of Computer Science and Engineering, The Hong Kong University of Science and Technology, Hong Kong, China. E-mail:~dongz@ust.hk.}
\IEEEcompsocthanksitem {S. He is with the School of Computing and Information Systems, Singapore Management University, Singapore. E-mail:~shengfenghe@smu.edu.sg.}
\IEEEcompsocthanksitem {Z. Li and J. Tang are the corresponding authors.}
}
}
% --------------------------------------------
\markboth{IEEE Transactions on Pattern Analysis and Machine Intelligence}{Tang \MakeLowercase{\textit{et al.}}: Divide-and-Conquer: Confluent Triple-Flow Network for RGB-T Salient Object Detection}
% --------------------------------------------	
\IEEEtitleabstractindextext{
\begin{abstract}
\justifying
RGB-Thermal Salient Object Detection (RGB-T SOD) aims to pinpoint prominent objects within aligned pairs of visible and thermal infrared images. 
A key challenge lies in bridging the inherent disparities between RGB and Thermal modalities for effective saliency map prediction.
Traditional encoder-decoder architectures, while designed for cross-modality feature interactions, may not have adequately considered the robustness against noise originating from defective modalities, thereby leading to suboptimal performance in complex scenarios. Inspired by hierarchical human visual systems, we propose the \textsc{ConTriNet}, a robust Confluent Triple-Flow Network employing a \textit{``Divide-and-Conquer"} strategy. This framework utilizes a unified encoder with specialized decoders, each addressing different subtasks of exploring modality-specific and modality-complementary information for RGB-T SOD, thereby enhancing the final saliency map prediction. 
Specifically, \textsc{ConTriNet} comprises three flows: two modality-specific flows explore cues from RGB and Thermal modalities, and a third modality-complementary flow integrates cues from both modalities. \textsc{ConTriNet} presents several notable advantages. It incorporates a \textit{Modality-induced Feature Modulator (MFM)} in the modality-shared union encoder to minimize inter-modality discrepancies and mitigate the impact of defective samples. Additionally, a foundational \textit{Residual Atrous Spatial Pyramid Module (RASPM)} in the separated flows enlarges the receptive field, allowing for the capture of multi-scale contextual information. Furthermore, a \textit{Modality-aware Dynamic Aggregation Module (MDAM)} in the modality-complementary flow dynamically aggregates saliency-related cues from both modality-specific flows. Leveraging the proposed parallel triple-flow framework, we further refine saliency maps derived from different flows through a \textit{flow-cooperative fusion strategy}, yielding a high-quality, full-resolution saliency map for the final prediction. To evaluate the robustness and stability of our approach, we collect a comprehensive RGB-T SOD benchmark, \textbf{VT-IMAG}, covering various real-world challenging scenarios. 
Extensive experiments on public benchmarks and our VT-IMAG dataset demonstrate that \textsc{ConTriNet} consistently outperforms state-of-the-art competitors in both common and challenging scenarios, even when dealing with incomplete modality data. The code and VT-IMAG will be available at:~\url{https://cser-tang-hao.github.io/contrinet.html}.
\end{abstract}
%
% --------------------------------------------
}
% --------------------------------------------
\maketitle
\IEEEdisplaynotcompsoctitleabstractindextext
\IEEEpeerreviewmaketitle
% --------------------------------------------	
\section{Introduction}
\label{section1}
% -----------------------------------------
\begin{figure*}[tbh!]
\centering
\includegraphics[width=0.97\linewidth]{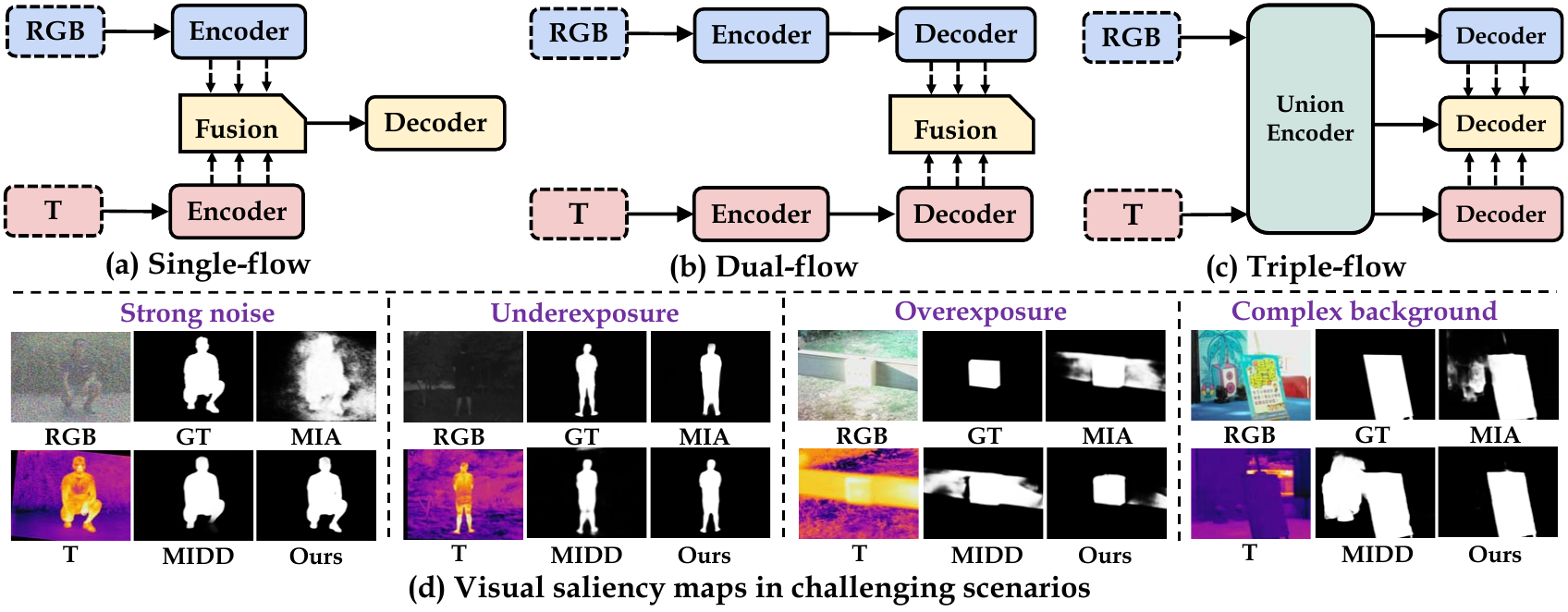}
\caption{Comparisons of established RGB-T SOD network architectures (\emph{i.e.,} (a) single-flow and (b) dual-flow) for RGB-T SOD and the proposed one in (c). Our proposed triple-flow paradigm adopts a \textit{divide-and-conquer} strategy dedicated to the deep exploration of modality-specific cues while the effective fusion of modality-complementary information, thus dealing with various challenging scenarios well (see (d)). MIA~\cite{liang2022multi} and MIDD~\cite{MIDD_21} correspond to the representative methods of (a) and (b), respectively.}%
\label{fig:first_image}
\vspace{-2mm}
\end{figure*}
% -----------------------------------------
\IEEEPARstart{S}{alient} Object Detection (SOD), a fundamental research task in various fields (\eg~computer vision, computer graphics, and robotics), aims to pinpoint pixel-level objects or regions that attract human visual attention within an image. Over recent years, SOD has been applied successfully in numerous downstream domains~\cite{GuoZ10, WangSYP18, ZhangLWLWL20, TangLLJTL24}. Particularly, significant progress has been made in the RGB-based SOD tasks~\cite{BorjiCHJL19,wang2021salient}. However, despite the rich texture and color information provided by the RGB modality, it lacks robustness and is easily affected by practical environmental factors. As shown in Figure~\ref{fig:first_image}, RGB-based methods struggle to achieve reliable SOD results in challenging real-world scenarios, primarily due to low image quality caused by poor lighting conditions, such as strong noise, underexposure, overexposure, and complex background. To address these limitations, various efforts~\cite{FanLZZC21, ZhouFCSS21, MIDD_21} have been made to introduce supplementary modalities that compensate for the shortcomings of the RGB modality, with a focus on RGB-Depth (RGB-D) and RGB-Thermal (RGB-T) modalities.

In RGB-D SOD, depth maps are crucial in estimating the distance between objects and the camera within a scene. This spatial information enhances the performance of SOD by providing crucial position and structural details. However, depth maps can be easily affected by adverse conditions such as poor illumination and bad weather, limiting their practical applications in real-world scenarios~\cite{WuGTAPD22, FuFJZSZ22, ZhangJ0FZ21}. On the other hand, the thermal modality, unlike the depth map, is insensitive to the surroundings and can effectively highlight the contour structure of salient objects. Nonetheless, it fails to capture the fine texture and details. Specifically, thermal infrared cameras capture the thermal radiation emitted by objects with temperatures above absolute zero, offering valuable object information even in challenging lighting conditions (see Figure~\ref{fig:first_image}).
Consequently, the combination of RGB and thermal modalities presents an ideal solution for SOD in complex scenes. However, directly applying well-established RGB-D SOD models to the RGB-T SOD task fails to yield satisfactory results due to disparate data characteristics. This work specializes in the RGB-T SOD task and aims to exploit the significance of thermal radiation emitted by objects and RGB-modality information to pinpoint prominent objects in diverse challenging scenes.

Compared to the RGB modality, which contains rich color and texture detail information, the thermal modality has the advantage of providing strong contrast information on temperature variations~\cite{MTMR_18, SunZL19, ADF_20}. This observation has prompted extensive investigations~\cite{CSRN_21, ECFFNet_21, MIDD_21} into the interaction and fusion of cross-modal features, while neglecting the intrinsic modality-specific information. Consequently, a pertinent question arises: \textbf{\textit{Is it necessary to design a specialized paradigm for RGB-T SOD that exploits the distinctive characteristics of each modality?}} This query is motivated by the fact that challenging scenes, as illustrated at the bottom of Figure.~\ref{fig:first_image}, often expose the limitations of both RGB and thermal modalities in providing discriminative cues for salient objects. Consequently, it leads to the contamination of aggregated features and yields unsatisfactory SOD results. Significantly, the challenges arising from defective modalities, caused by factors such as strong noise, thermal crossover, and adverse weather conditions, in complex real-world scenarios or imaging processes, have been largely overlooked. Addressing these complexities raises another challenging question: \textbf{\textit{How can we develop a robust RGB-T SOD model suitable for practical applications in complex scenarios?}} Consequently, an effective design for an RGB-T SOD model necessitates a comprehensive consideration of robustness in complex scenarios, ensuring efficient and accurate SOD even under less-than-ideal or dynamically changing environmental conditions.

Transitioning from the aforementioned challenges, it can be observed that existing  RGB-T SOD methods exploit RGB- and thermal-modality information using dual encoders to extract individual features. These models broadly fall into two paradigms: the single-flow and the dual-flow, based on the different decoding strategies. Figure~\ref{fig:first_image}(a) illustrates the \textit{single-flow} paradigm (\emph{e.g.,~}\cite{ECFFNet_21, CSRN_21,liang2022multi}), in which cross-modality fusion for multi-scale features is performed during the encoding phase, followed by a singular decoder to predict the final saliency map. In contrast, Figure~\ref{fig:first_image}(b) showcases the \textit{dual-flow} paradigm adopted by methods such as MIDD~\cite{MIDD_21} and CGFNet~\cite{CGFNet_21}. Here, twin parallel decoders reconstruct salient-related features from each modality, and these hierarchical features are then fused to generate the final saliency map. Despite the remarkable performance achieved by these existing methods, several challenges remain unresolved. 
Firstly, the single-flow architecture grapples with defective inputs as it primarily focuses on fusing modality-complementary features, often leading to unreliable cross-modality feature aggregation due to the lack of modality-specific supervision, as shown in the saliency maps of MIA~\cite{liang2022multi} in Figure~\ref{fig:first_image}(d). Secondly, various unknown factors in cluttered scenarios may cause the salient objects to be defective or dominated by a single modality. The dual-flow architecture, although incorporating features from both modalities, often falls short in addressing the modality discrepancy and mitigating the negative impact of a defective modality. As demonstrated in the last group of visual results in Figure~\ref{fig:first_image}(d), both MIA~\cite{liang2022multi} and MIDD~\cite{MIDD_21} struggle to accurately discriminate salient objects from similar backgrounds. Inspired by the resilience and adaptability inherent in human perception, this work proposes a specialized  RGB-T SOD framework that adopts a \textbf{\textit{``Divide-and-Conquer''}} strategy to deeply explore modality-specific information and effectively integrate modality-complementary information.

In the human visual perception, the brain effectively processes multi-sensory input through parallel flows~\cite{stein1993merging}, where each flow decodes distinctive aspects of information, allowing for compensation whenever one sensory modality is compromised, thus maintaining holistic environmental perception~\cite{goodale1992, muggleton2011cognitive}. As illustrated in Figure~\ref{fig:first_image}(c), we introduce the Confluent Triple-Flow Network (\textsc{ConTriNet}), inspired by this cognitive process, with the objective of enhancing the robustness of RGB-T SOD by decomposing the task into two subtasks: modality-specific information mining and modality-complementary information integration. 
Firstly, we employ a modality-shared union encoder to derive intricate multi-layer features from dual-modality inputs via a shared backbone--a departure from traditional dual encoders. In this unified encoder, our proposed \textit{Modality-induced Feature Modulator (MFM)} refines and merges two groups of multi-level features, minimizing redundant information and modality discrepancies while preserving cross-modality consistency in each feature scale. The essence of the \textbf{\textit{``Divide-and-Conquer''}} strategy lies in the disentangled flows, which independently explore modality-specific and modality-complementary information, facilitating mutual compensation in cases of modality deficiency. 
Specifically, our \textit{Modality-aware Dynamic Aggregation Module (MDAM)} integrates salient-related cues from modality-specific flows into the modality-complementary flow in a dynamic weighting manner, selectively controlling the contribution of hierarchical complementary cues for accurate salient object localization and bias reduction. 
The triple flows of the \textsc{ConTriNet} incorporate our \textit{Residual Atrous Spatial Pyramid Module (RASPM)} as a fundamental component, which enhances the effective receptive field and captures contextual information at multiple scales. To promote consistent salient features, label supervision rectifies the three parallel decoders of the triple-flow network. Subsequently, a \textit{flow-cooperative fusion strategy} is employed to generate a comprehensive saliency map, mimicking integrated perception in humans. This approach effectively highlights salient objects and mitigates the negative impact of defective modalities in challenging scenarios, thereby enhancing robustness and precision in RGB-T SOD. To assess the robustness of RGB-T SOD models, we introduce a self-collected \textbf{VT-IMAG} dataset as a benchmark testbed, driving efforts towards robust architectural exploration. Extensive experiments demonstrate that the proposed \textsc{ConTriNet} outperforms state-of-the-art approaches by a significant margin on three public benchmarks (\ie~VT821~\cite{MTMR_18}, VT1000~\cite{SGDL_20}, and VT5000~\cite{ADF_20}), as well as our challenging VT-IMAG benchmark.
The main contributions of this work are summarized as follows:
\begin{itemize}
\item A novel \textit{``Divide-and-Conquer''} strategy is introduced to address the weak robustness of current RGB-T SOD models. Within this context, we propose the \textsc{ConTriNet}, specified for effective fusing modality-complementary cues and deeply mining modality-specific cues.
\item An MFM is embedded in our union encoder to effectively narrow modality discrepancies and filter redundant or potentially disruptive information. Concurrently, an MDAM has been engineered into our modality-complementary flow to dynamically prioritize salient-related features from modality-specific flows while mitigating salient bias.
\item We present the RASPM, a cornerstone for our parallel flows that offers a larger yet compact receptive field. Stemming from this, we further introduce a flow-cooperative fusion strategy, aiming for a refined, comprehensive saliency map prediction.
\item A comprehensive benchmark with various challenging scenarios is contributed to the field as a new testbed for the faithful RGB-T SOD evaluations. 
\end{itemize}

\section{Related Work}
\label{section2}
% -----------------------------------------
\begin{figure*}[t!]
\centering
\begin{tabular}{c}
\includegraphics[width=\linewidth]{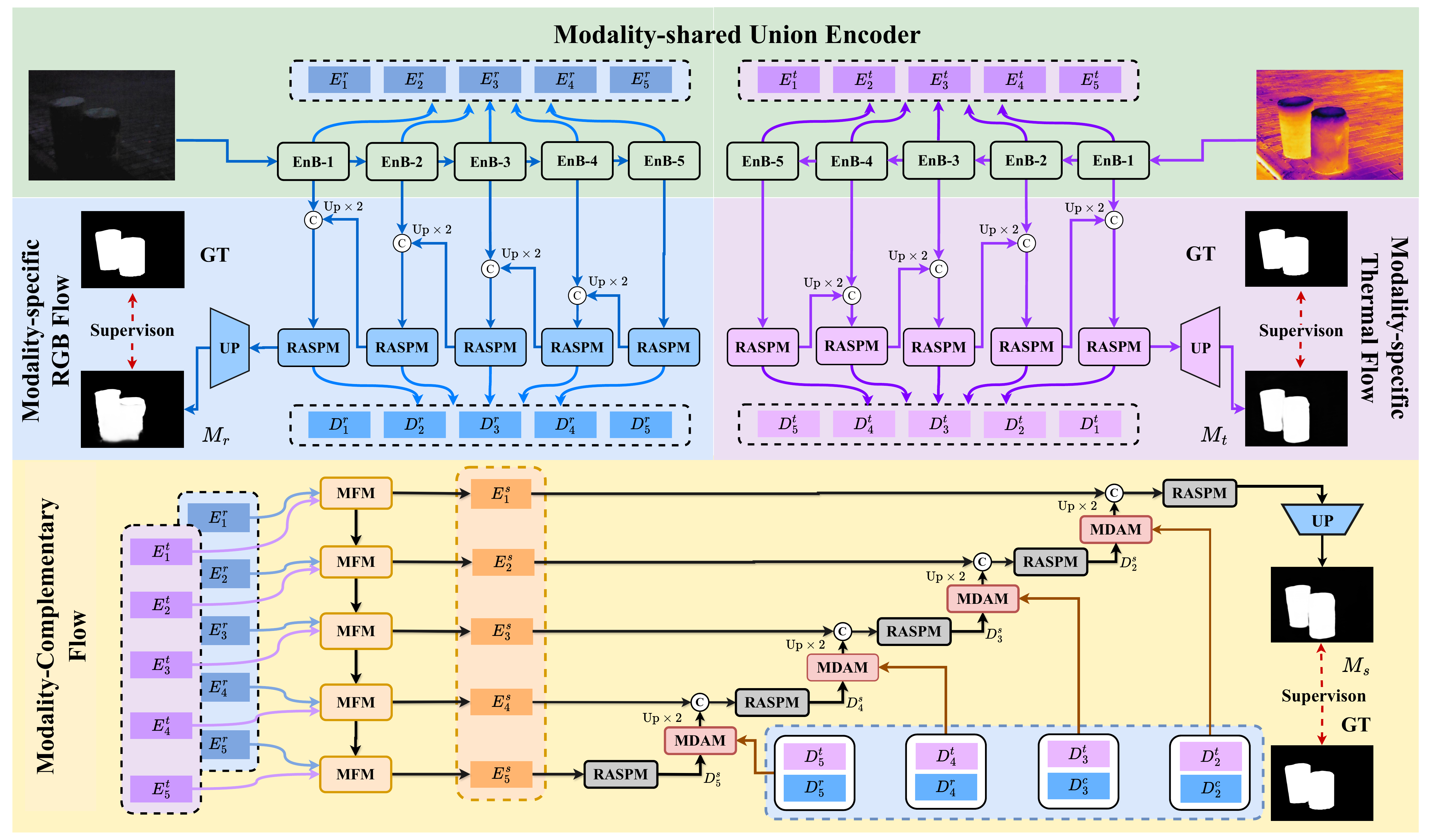} \\
\end{tabular}
\caption{An overview of the proposed Confluent Triple-Flow Network (\textsc{ConTriNet}), which adopts an efficient \textbf{``Divide-and-Conquer''} strategy, is presented. \textsc{ConTriNet} comprises three main flows: a modality-complementary flow that predicts a modality-complementary saliency map, and two modality-specific flows that predict RGB- and Thermal-specific saliency maps, respectively. The union encoder of two modality-specific flows shares parameters and the overall framework can be trained end-to-end.}%
\label{fig:network}
\vspace{-2mm}
\end{figure*}
% -----------------------------------------

% -----------------------------------------
\subsection{RGB Salient Object Detection}
% -----------------------------------------
Traditional salient object detection (SOD) methods primarily rely on hand-crafted features (\eg,~color, texture, and edge features)~\cite{LiLZRY13,ChengMHTH15,WangJYCHZ17} and intrinsic cues (\eg,~color contrast and edge density)~\cite{JiangD13,ZhuLW014,ZhouYYZH15} for predicting regional saliency scores. However, these approaches suffer from limitations in terms of generalization and effectiveness due to the labor-intensive and time-consuming steps of saliency map extraction. In recent years, the emergence of deep learning techniques has led to significant advancements in computer vision and multimedia. Consequently, contemporary deep learning-based SOD methods extensively employ convolutional neural networks (CNNs)~\cite{SimonyanZ14a,HeZRS16,HouCHBTT17,ZhangWQLW18,GaoCZZYT21} to generate pixel-wise predictions utilizing various saliency models, achieving remarkable results. For example, Hou~\emph{et al.}~\cite{HouCHBTT17} augmented location information in higher-level features by incorporating additional short connections from deeper to shallower side outputs, enhancing accuracy. Similarly, Zhang~\emph{et al.}~\cite{ZhangWQLW18} integrated pyramid attention mechanisms within a U-shape network to recurrently enhance the representation ability of multi-scale features. Recent saliency models~\cite{WuSH19,f3net_20,ChenXCH20,WeiWWSH020,ZhangLPYL21} have made significant progress by fusing multi-scale context features through various refinement strategies. For instance, Pang~\emph{et al.}~\cite{PangZZL20} aggregated and interacted features from adjacent layers to enhance multi-scale contextual features. Zhang~\emph{et al.}~\cite{ZhangLPYL21} proposed a novel NAS-based framework to automatically search the optimal fusion strategy for multi-scale features, eliminating the need for manual operations. Additional insights into RGB SOD methodologies can be found in recent review articles~\cite{BorjiCHJL19,wang2021salient}. 

However, practical applications pose challenges including strong noise, low illumination, and complex backgrounds. For instance, traditional RGB-based SOD methods may fail under low illumination conditions as they struggle to accurately capture salient objects. These challenges necessitate the exploration of auxiliary modalities, herein referred to as RGB-X SOD (\eg,~RGB-D and RGB-T). This work focuses on RGB-T SOD, aiming to pinpoint prominent objects or regions that are common to both RGB and thermal modalities. The incorporation of thermal modality not only alleviates the aforementioned challenges but also improves the robustness of our model in adverse environments, facilitating more accurate and reliable salient object detection in a wide range of real-world scenarios.

% ---------------------------------------
\subsection{RGB-X Salient Object Detection}
The affordability and portability of sensors have facilitated the acquisition of multi-modality data, enabling its utilization in complex scenarios to address the inherent challenges of RGB SOD. The adoption of RGB-X SOD, particularly RGB-D SOD, has gained momentum due to the invaluable depth information provided by depth maps generated by depth sensors, driving the advancement of RGB-D SOD methods~\cite{CDINet_21, ZhangJ0FZ21, DCMF_22}, with more insights available in recent literature reviews~\cite{FanLZZC21, ZhouFCSS21}. Although depth maps contain rich spatial and structural information, they are vulnerable to disturbances in poorly illuminated and occluded environments.
In parallel, RGB-T SOD combines RGB and thermal infrared (T) maps, capitalizing on the complementary advantages of temperature cues. The robustness of thermal information in challenging environments, compared to depth information, highlights the need to explore RGB-T SOD. 
Despite the growing interest in RGB-T SOD within the research community, a gap still exists in investigating robust salient patterns to tackle challenging scenes.

Traditional RGB-T SOD methods typically rely on hand-crafted features. The first RGB-T SOD dataset, {VT821}, was established in MTMR by Wang~\emph{et al.}~\cite{MTMR_18}, where a multi-task manifold ranking algorithm was proposed to capture the saliency cues. Subsequent efforts include the work of Tu~\emph{et al.}~\cite{M3S-NIR_19}, who introduced an intermediate variable for inferring optimal seed nodes to perform multi-modal multi-scale graph-based manifold ranking. Additionally, Tu~\emph{et al.}~\cite{SGDL_20} proposed a collaborative graph learning method that utilizes hierarchical deep features to jointly learn graph affinity and node saliency. They also constructed a more challenging dataset, {VT1000}. However, these attempts mainly rely on machine learning techniques (\emph{e.g.}, SVM~\cite{MaSMD017}, ranking models~\cite{MTMR_18, M3S-NIR_19}, and graph learning~\cite{SGDL_20}), struggling to achieve satisfactory performance due to the limited semantic representation capacity of hand-crafted features. 
Deep learning-based methods~\cite{FMCF_20,ADF_20,MIDD_21,CGFNet_21,liang2022multi} have made significant progress in RGB-T SOD with the application of deep neural networks. Zhang~\emph{et al.}~\cite{FMCF_20,ZhangXHZH21} regarded RGB-T SOD as a features fusion problem and proposed two CNNs-based models to fuse the features from multi-scale, multi-level, and multi-modality perspectives. Concurrently, Tu~\emph{et al.}~\cite{ADF_20} proposed a multi-interactive dual-decoder network for the aggregation of cross-modality features and global contexts, and they created a large-scale RGB-T dataset named VT5000. Furthermore, Zhou~\emph{et al.}~\cite{ECFFNet_21} proposed a bilateral reversal fusion approach that effectively and consistently fuses cross-modality features of foreground and background information. Huo~\emph{et al.}~\cite{CSRN_21} introduced a context-guided cross-modality fusion module to exploit complementary cues and used a stacked refinement network to enhance salient regions. Moreover, Tu~\emph{et al.}~\cite{MIDD_21} proposed a multi-interactive dual-stream decoder to capture multi-type cues from the two modalities, while Wang~\emph{et al.}~\cite{CGFNet_21} achieved a more comprehensive exploration of single-modality information and a more effective integration of cross-modality information through the proposed cross-guided fusion module. 
Different from them, we focus on fully
exploiting the respective characteristics of RGB and thermal
modalities for the robust RGB-T SOD. \textsc{ConTriNet} is proposed
to explore saliency-oriented modality-specifc and modality-complementary information in a more reasonable way, which
performs well on a variety of challenging benchmarks.

% ##################################
\subsection{Our Key Differentiators}
Following the ``\textit{Divide-and-Conquer}'' strategy, our work introduces \textsc{ConTriNet}, which systematically explores both modality-specific and modality-complementary information. This approach significantly diverges from recent three-stream frameworks such as SPNet~\cite{SPNet_21} and CIR-Net~\cite{CongLZLCHZ22}, which are primarily tailored for RGB-D SOD tasks. Below, we outline the key differentiators of \textsc{ConTriNet}, which are crucial in handling the unique challenges of RGB-T SOD tasks, providing both conceptual and practical advancements over existing methods.
(1) \textbf{Functional Strategy:} 
Unlike conventional frameworks that often treat additional modalities as supplementary, \textsc{ConTriNet} assigns equal importance to both RGB and thermal modalities from the onset. This ensures comprehensive utilization of both modalities, enabling dynamic interactions and holistic integration. This strategy goes beyond mere structural design and serves as a functional approach to maximize the strengths and mitigate the weaknesses of each modality through dedicated processing paths.
(2) \textbf{Architectural Design:} 
\textsc{ConTriNet} significantly deviates from the common practice of using separate encoders for each modality followed by a shared network for feature fusion. Instead, it employs a unified encoder that works in conjunction with three specialized decoders, each dedicated to a specific aspect of the task. This design reduces complexity and boosts efficiency, aligning  with the \textit{``Divide-and-Conquer''} philosophy to improve feature integration and processing.
(3) \textbf{Robust Perception:} 
\textsc{ConTriNet} departs from the traditional reliance on a single decoder for final saliency prediction, a characteristic shared by frameworks such as SPNet~\cite{SPNet_21} and CIR-Net~\cite{CongLZLCHZ22}, It introduces a triple-flow architecture where each flow addresses specific subtasks to capture modality-specific or modality-complementary cues, operating in parallel. \textsc{ConTriNet} also incorporates a \textit{flow-cooperative fusion strategy} for inference, ensuring more precise and robust predictions, particularly effective in handling diverse and challenging scenarios that may overwhelm single-decoder designs.
(4) \textbf{Tailored Modules:}
\textsc{ConTriNet} incorporates tailored modules (\ie, MFM, RASPM, and MDAM) specifically designed for RGB-T SOD tasks. These modules are critical for adapting the processing strategy to the unique challenges posed by thermal data, thereby enhancing both the efficacy and efficiency of feature integration.
In summary, while there are certain similarities with SPNet~\cite{SPNet_21} and CIR-Net~\cite{CongLZLCHZ22}, \textsc{ConTriNet} outperforms them in the RGB-T SOD task by effectively addressing the unique challenges presented by thermal and RGB data, thereby facilitating accurate and robust prediction in a wide range of real-world scenarios.

\section{METHODOLOGY}
\label{section3}
% ------------------------------------
\subsection{Overview}
% ------------------------------------
\myparagraph{Background.}
The \textit{``Divide-and-Conquer''} strategy~\cite{Algorithms} is a well-established computational principle that decomposes a complex problem into smaller, manageable sub-problems. Each sub-problem is then handled individually, and the solutions are merged to resolve the original problem. This strategy simplifies problem-solving by breaking down complex challenges into manageable tasks, often resulting in more efficient and optimized solutions as specialized attention can be given to individual sub-problems. 
Remarkably, the human visual system employs this strategy naturally. Instead of processing an entire scene simultaneously, the brain dissects various aspects of visual stimuli~\cite{goodale1992}. Different components of the visual pathways decode specific elements such as color, shape, or movement~\cite{livingstone1988}. These separately processed cues are subsequently integrated to ensure a comprehensive perception of our surroundings, even in cases where one sensory modality is impaired.

% ------------------------------------
\myparagraph{Motivation.}
% ------------------------------------
Inspired by the adaptive \textit{``Divide and Conquer''} capability of the human visual system, where different components of the visual pathways decode specific elements and synthesize them for comprehensive perception, our objective is to improve the accuracy and robustness of RGB-T SOD. Similar to how the brain systematically processes diverse sensory inputs to comprehend a scene, we aim to investigate the intrinsic characteristics unique to each modality in RGB-T SOD. 
Specifically, RGB images provide detailed structural and texture information but are susceptible to environmental interference. Conversely, thermal images prioritize structure delineation at the expense of texture detail, yet exhibit higher resilience to environmental variations, such as lighting conditions. To accurately locate visually prominent objects in complex scenarios, we propose two core principles: \textit{deep mining of modality-specific information} and \textit{effective integration of modality-complementary information}. The former aims to suppress noisy or misleading inputs from each modality, while the latter ensures that the saliency predictions are not overly dominated by a single modality, thus fostering a more balanced and robust model. By adhering to these principles, we aim to significantly enhance the quality of SOD performance. Consequently, this work introduces a confluent triple-flow network, incorporating our ``Divide and Conquer'' strategy by addressing modality-specific and modality-complementary subtasks within a parallel triple-flow framework. This design enables more accurate and resilient perception for RGB-T SOD in various challenging scenarios, leveraging the strengths of each modality while mitigating their individual weaknesses.

% ------------------------------------
\myparagraph{Architecture.}
% ------------------------------------
Figure~\ref{fig:network} illustrates the architecture of the proposed \textsc{ConTriNet}, which aims to decompose the final prediction of the RGB-T SOD task into modality-specific and modality-complementary subtasks. This decomposition allows for leveraging the unique strengths of each modality to improve the accuracy and adaptability of the model. Specifically, the main architecture of \textsc{ConTriNet} are: 
(\romannumeral1) A union encoder with a shared backbone for extracting multi-scale low- and high-level features from dual-modality inputs, coupled with a Modality-induced Feature Modulator~(MFM) to reduce modality discrepancies and seamlessly integrate features from both two modalities. 
(\romannumeral2) Two modality-specific flows dedicated to refining multi-scale features specific to each modality, aiming to predict saliency maps for RGB and thermal domains respectively. 
(\romannumeral3) A modality-complementary flow for dynamically integrating saliency-related cues to generate a modality-complementary saliency map. Within this flow, a Modality-aware Dynamic Aggregation Module~(MDAM) is introduced to merge discriminative cues from the two modality-specific flows, aiding in the inference of more consistent salient regions. 
(\romannumeral4) A Residual Atrous Spatial Pyramid Module~(RASPM) is deployed within these flows to provide a larger yet compact receptive field. Meanwhile, a flow-cooperative fusion strategy to combine the separately predicted results of these flows, producing a comprehensive saliency map. 
Following the Res2Net-50~\cite{GaoCZZYT21} architecture, the modality-shared union encoder extracts hierarchical features from the paired RGB and thermal images, denoted as $E_{i}^{r}$ and $E_{i}^{t}$, respectively, where $i\in \{1,2,\ldots, 5\}$ indicates the feature level.

% --------------------------------
\begin{figure}[t!]
	\centering
	\begin{minipage}[t]{\linewidth}
        \centering
	\includegraphics[width=\linewidth]{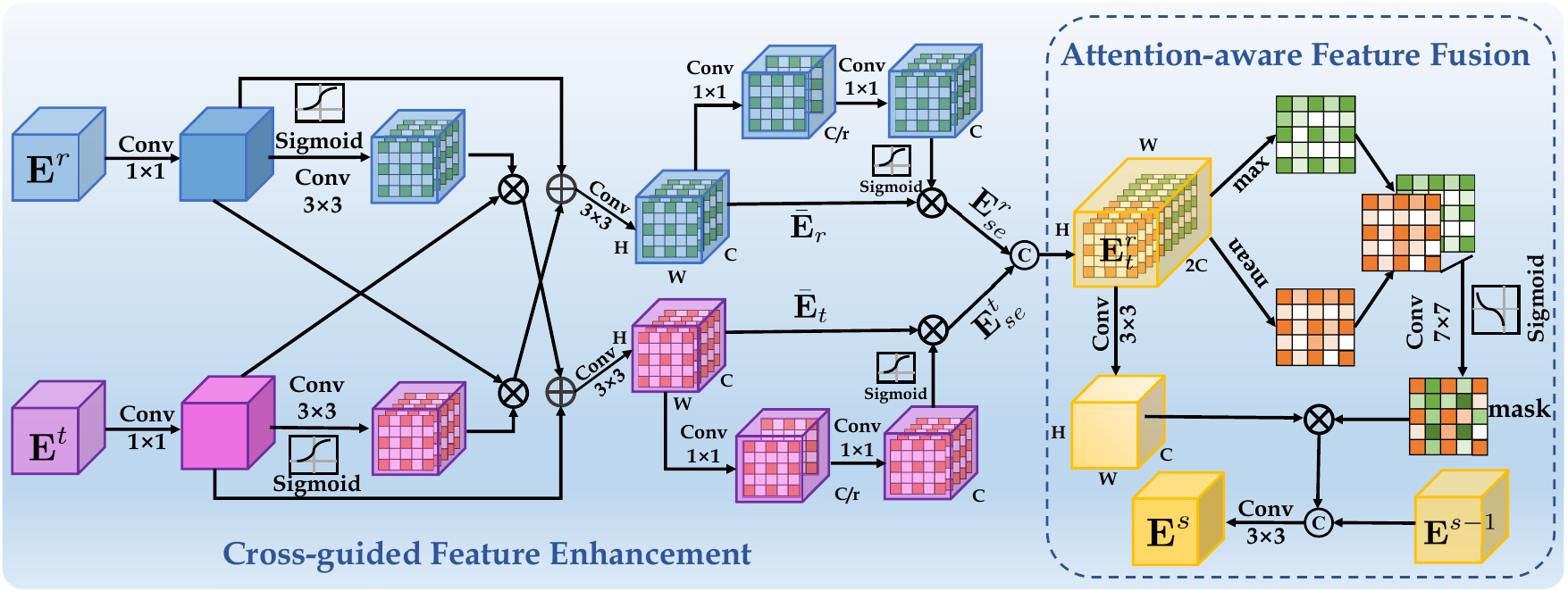}
        \caption{Structure diagram of the Modality-induced Feature Modulator (MFM).}
        \label{fig:AGFM}
        \end{minipage}\vspace{2mm}
        \begin{minipage}[t]{\linewidth}
        \centering
        \includegraphics[width=\linewidth]{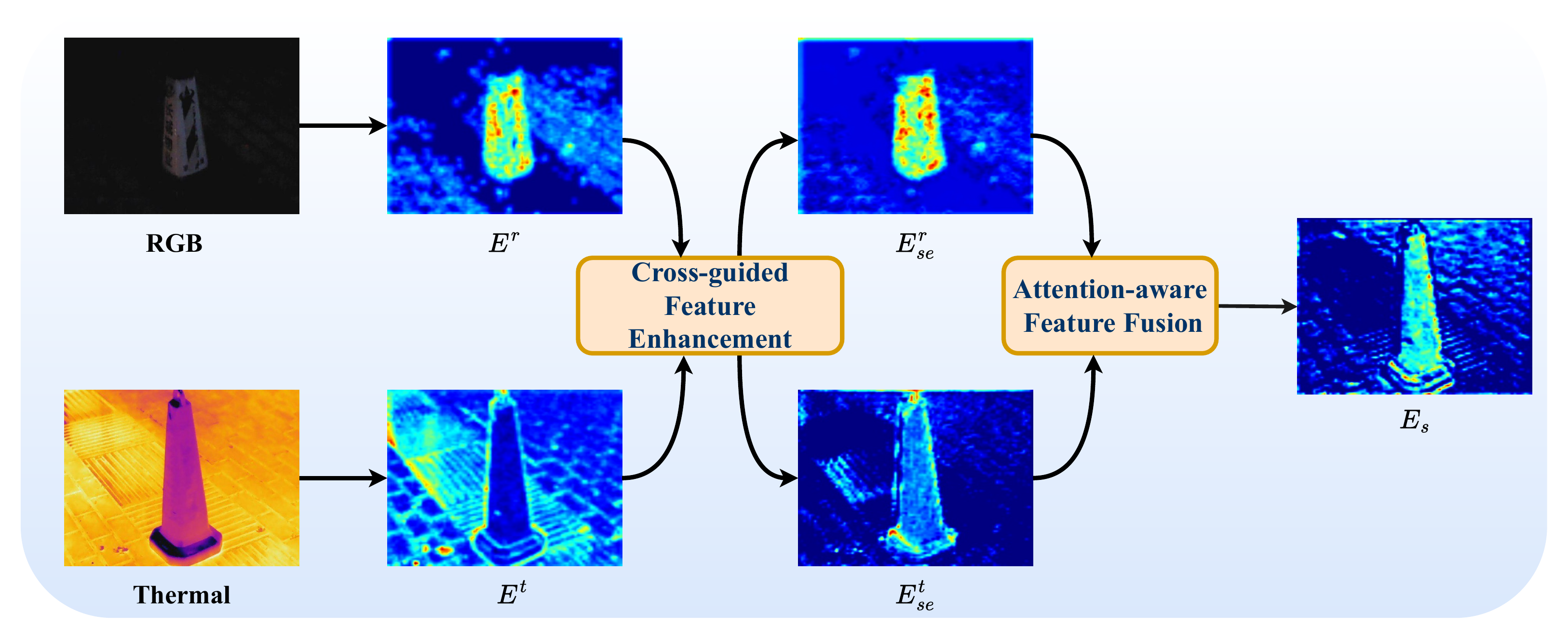}
        \caption{Visualization for the feature evolution in the modality-induced feature modulator.}
        \label{fig:vis_feature}
	\end{minipage}
	\vspace{-2mm}
\end{figure}
% --------------------------------

\subsection{Modality-Induced Feature Modulator}\label{MFM}
% ------------------------------------------
In order to address two inevitable challenges encountered in RGB-T SOD under extremely complex environments, namely the obstacle of merging features from different modalities due to intrinsic modality discrepancy and the problem of noise interference caused by defective inputs, we propose the incorporation of a Modality-induced Feature Modulator (MFM) within the modality-shared union encoder. This integration allows for profound refinement and efficient fusion of modality-specific patterns learned in each feature scale. As illustrated in Figure~\ref{fig:AGFM}, MFM comprises two stages: cross-guided feature enhancement and attention-aware feature fusion.

\textbf{Cross-guided Feature Enhancement.}
To mitigate the discrepancy among various modalities while preserving their discriminative capabilities, we propose a cross-guided feature enhancement module. Taking $\{E_{i}^{r}\}_{i=1}^5$ and $\{E_{i}^{t}\}_{i=1}^5$ as inputs, which are derived by the modality-shared union encoder, we first adopt a group of general operations (\emph{i.e.,} $\left\{\textit{Conv}_{1\times1}\left(\cdot\right), \textit{Conv}_{3\times3}\left(\cdot\right), \textit{Sigmoid}\left(\cdot\right) \right\}$) to normalize them as $w_{i}^{r}$ and $w_{i}^{t}$, so as to model cross-modality long-range interdependency by weighting the RGB and thermal features, thereby determining the regions that require feature complementation. Moreover, residual connections are employed to preserve the original information of each modality. 
Hence, the enhanced RGB and thermal features, denoted as $\bar{E}_{r}$ and $\bar{E}_{t}$, are formulated as
\begin{equation}	
	\begin{split}
		\bar{E}_{r} = E^{r} + E^{r} \otimes w^{t},\\
		\bar{E}_{t} = E^{t} + E^{t} \otimes w^{r}.
	\end{split} 
\end{equation}
To emphasize the salient content of modality-specific features, it is necessary to recalibrate them based on their characteristics and their contributions in encoding discriminative information. For this purpose, we employ the \textit{Squeeze-Excitation} (SE) operation along the channel dimension, which not only selects representative channels but also suppresses noise between modalities to enhance denoising at the feature level. The SE operation is implemented through two sequential $1\times 1$ convolution operations, designed to first reduce and then expand the channel dimensions (\ie,~\emph{$c\mapsto c/r\mapsto c$}), enabling a dynamic recalibration of feature channels.
Mathematically, this denoising and recalibration operation is structured as follows:
\begin{equation}	
	\begin{split}
		E_{se}^{r} =\mathcal{S}( \textit{C}_{se}\left(\textit{Conv}_{3\times3}\left(\bar{E}_{r}\right)\right)) \otimes \bar{E}_{r},\\
		E_{se}^{t} = \mathcal{S}( \textit{C}_{se}\left(\textit{Conv}_{3\times3}\left(\bar{E}_{t}\right)\right)) \otimes \bar{E}_{t},
	\end{split}
\end{equation}
where $\textit{C}_{se}\left(\cdot\right)$ denotes the SE operation, $\otimes$ denotes element-wise multiplication, and $\mathcal{S}(\cdot)$ represents the \textit{Sigmoid} function. The normalized outputs of $\textit{C}_{se}\left(\cdot\right)$ can be viewed as channel attention maps that reflect the importance of different channels in RGB and thermal features. In summary, this stage enhances the saliency representations of RGB and thermal features, acting as a pre-processing step for regional feature compensation.

\textbf{Attention-aware Feature Fusion.}
Given the improved cues from different modalities~(\emph{e.g.,} $E_{se}^{r}$ and $E_{se}^{t}$), we logically enhance the compatibility of cross-modality features and achieve consistent fusion. As both RGB and thermal features provide significant discriminative information for SOD, we first utilize a feasible concatenation feature aggregation strategy to integrate $E_{se}^{r}$ and $E_{se}^{t}$, resulting in the preliminary fusion feature $E_{t}^{r}$. However, the simple concatenation operation fails to consider the independence and inconsistency of modality-specific information in RGB and thermal images. To address this issue, we draw inspiration from the spatial attention mechanism~\cite{CBAM_18}, where we introduce global average pooling $\mathbf{A}(\cdot)$ and max pooling $\mathbf{M}(\cdot)$ operations along the channel axis simultaneously to the feature $E_{t}^{r}$ in order to calculate spatial statistics for locating salient objects. We then concatenate the parallel pooling results to enhance the spatial structure features. As depicted within the dashed box on the right side of Figure~\ref{fig:AGFM}, the modulated output $E^{s}$ is obtained through 
\begin{equation}
\resizebox{0.9\hsize}{!}{$
E^{s}=\textit{Conv}_{3 \times 3}\left(E_{t}^{r}\right) \otimes \mathcal{S}\left(\textit{Conv}_{7 \times 7}\left(\left[\mathbf{A}\left(E_{t}^{r}\right) ; \mathbf{M}\left(E_{t}^{r}\right)\right]\right)\right),$}
\end{equation}
where $\mathcal{S}(\cdot)$ denotes the \textit{Sigmoid} activation function, and $\left[\cdot~;~\cdot\right]$ denotes the concatenation operation. 

Figure~\ref{fig:network} illustrates the construction of the MFM for achieving the adaptive fusion of two groups of multi-level multimodal features from the modality-shared union encoder in a coarse-to-fine
manner. 
Specifically, at the $i$-\textit{th} feature level, the current output $E_{i}^{s}\in E^{s}$ is concatenated with the previous output $E_{i-1}^{s}\in E^{s}$ of the ($i$-1)-\textit{th} MFM. This concatenated feature is then passed through a convolutional layer with a kernel size of $3\times 3$, resulting in the final fused feature at the $i$-\textit{th} feature level. Importantly, the current modulated feature $E_{i}^{s}$ is propagated to the ($i$+1)-\textit{th} layer to explore and integrate cross-level and multi-scale cues. Only the current feature $E_{1}^{s}$ is fed into the $\textit{Conv}_{3\times 3}$ layer when $i=1$. 
Here, we deliberately select a challenging low-light scene with a reflective traffic cone as the salient object to showcase the efficacy of the proposed MFM. In this particular scenario, the visible region exhibits complementarity from a multi-modality perspective. The RGB and thermal features originating from the shallow layer of the union encoder, denoted as $E^{r}$ and $E^{t}$ respectively, are visualized in Figure~\ref{fig:vis_feature}. The enhanced features $E^{r}_{se}$ and $E^{t}_{se}$, obtained after the \textit{cross-guided feature enhancement} phase, effectively suppress noise in the background while further enhancing the activation values in the foreground. The complementarity is still evident in $E^{r}_{se}$ and $E^{t}_{se}$. Post the \textit{attention-aware feature fusion} phase,  the modality-complementary fused feature $E^{s}$ demonstrates a well-balanced activation, focusing on the entire traffic cone. This showcases the MFM's capability to leverage complementary information from modality-specific features and progressively mitigate the inherent disparity between modalities to construct the modality-complementary fused feature.

% ------------------------------------------
\subsection{Residual Atrous Spatial Pyramid Module}\label{RASPM}
Given the acquired hierarchical discriminative features $\{E^{r}, E^{t}, E^{s}\}$ derived from a modality-shared union encoder, our primary objective is to maximize their potential for enhancing the effectiveness of the multi-modality representations in generating precise saliency maps. 
It is widely recognized that both local and global semantic information play crucial roles in SOD tasks. The shallow layers of the network are responsible for learning local semantic information, while the global information depends on the receptive field size of the network. However, the modality-shared union encoder, which employs serial convolution operations, fails to capture abundant context information. Additionally, the narrow usage of $3\times 3$ convolutions hinders the expansion of the effective receptive field~\cite{LiSZT22}, thereby adversely affecting obtaining fine structures and clear boundaries within the predicted saliency map. To solve these problems, we propose a lightweight Residual Atrous Spatial Pyramid Module (RASPM) as the core component within our specialized triple flows to effectively capture compact contextual information from multiple receptive fields and feature scales.

\begin{figure}[t!]
	\centering
	\begin{tabular}{c}
	\includegraphics[width=\linewidth]{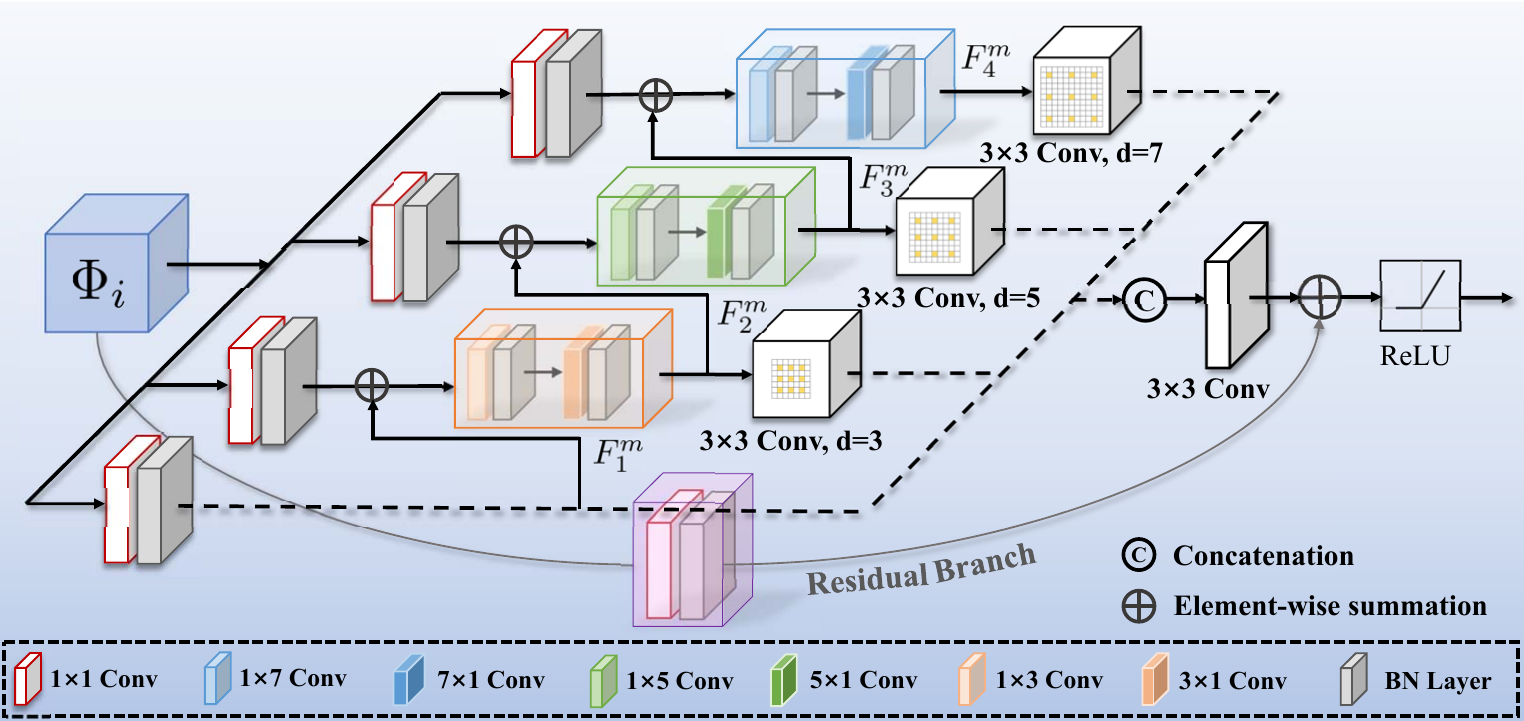} \\
	\end{tabular}
	\vspace{-1mm}
	\caption{Structure diagram of the Residual Atrous Spatial Pyramid Module (RASPM).}%
	\label{fig:RASPB}
	\vspace{-2mm}
\end{figure}
% ------------------------------------------

As illustrated in Figure~\ref{fig:RASPB}, the RASPM is composed of four parallel branches dedicated to capturing context features and a residual branch aimed at preserving the original features. Each branch corresponds to a specific feature scale. To be specific, a $1\times 1$ convolution layer is included in each branch to reduce the channel size. Additionally, the $k$-th branch (where $k\in \{2,3,4\}$) is equipped with a $1\times (2k-1)$ asymmetric convolution followed by a $(2k-1)\times 1$ asymmetric convolution in order to decrease the computational load. Given the correlation and potential mutual benefit of features with different scales, a cross-branch short-cut connection is introduced to integrate features from the bottom to the top and avoid information loss. For example, considering the $i$-\textit{th} RASPM of RGB-modality flow, its input is denoted as $\Phi_{i}$, and the intermediate features ${F}_{k}^{m}$ learned by the four branches are defined as
\begin{equation}
\resizebox{0.92\hsize}{!}{$
	\begin{split}
		{F}_{k}^{m} = \left\{\begin{array}{lc}
		\textit{Conv}_{1\times1}(\Phi_{i}) & k=1 \vspace{1ex} \\
		\textit{Conv}_{2k-1}^{1}(\textit{Conv}_{1\times1}(\Phi_{i}) +{F}_{k-1}^{m}) & k\in\{2,3,4\}
		\end{array}\right.
	\end{split},$}
\end{equation}
where $\textit{Conv}_{2k-1}^{1}$ presents the stacked convolution operations with a kernel size of $(2k-1)\times 1$ and $1\times (2k-1)$. In order to attain an expanded receptive field without sacrificing feature resolution, we incorporate atrous convolution operations with a dilation rate of $2k-1$ into the four parallel branches, forming an atrous spatial pyramid. Subsequently, the outputs of the four branches are concatenated and then processed by a $3\times3$ convolution to adaptively reweight them and reduce the channel size. Finally, the reconstructed features with rich context information are generated by a residual combination of multi-scale features and original features, defined as:
\begin{equation}	
\resizebox{0.92\hsize}{!}{$
	\begin{split}
		D_{i}^{m} = \textit{Conv}_{3\times3}([\textit{DConv}_{2k-1}({F}_{k}^{m})]_{k=1}^{4})+
		\textit{Conv}_{1\times1}(\Phi_{i}),
	\end{split}$}
\end{equation}
where $\textit{DConv}_{2k-1}$ denotes $3\times3$ convolution with a dilation rate of $2k-1$. It is worth noting that all convolutions in our RASPM are followed by batch normalization.

In this way, the multi-scale feature sets $E^{r}$ and $E^{t}$ are combined through the RGB- and thermal-modality flows in a layer-by-layer manner. Figure~\ref{fig:network} illustrates the acquisition of $D^{r/t}$ from the modality-specific RGB/Thermal flows as follows:
\begin{equation}	
\resizebox{0.92\hsize}{!}{$
	\begin{split}
		{D}_{i}^{r/t} = \left\{\begin{array}{lc}
		\textrm{RASPM}({E}_{i}^{r/t}) & i=5 \vspace{1ex} \\
		\textrm{RASPM}([\textit{UP}_{\times2}({D}_{i+1}^{r/t}));{E}_{i}^{r/t}]) & i\in\{1,2,3,4\}
		\end{array}\right.
	\end{split},$}
\end{equation}
where $\textit{UP}_{\times2}$ denotes a $2\times$ upsampling operation and $\left[\cdot~;~\cdot\right]$ denotes the concatenation operation. Subsequently, the obtained $D^{r}$ and $D^{t}$ are integrated into the modality-complementary flow to provide saliency-related information necessary for dynamic selection, which will be discussed in Section~\ref{MDRM}.

\begin{figure}[t!]
%	\vspace{-2mm}
	\centering
	\begin{tabular}{c}
	\includegraphics[width=\linewidth]{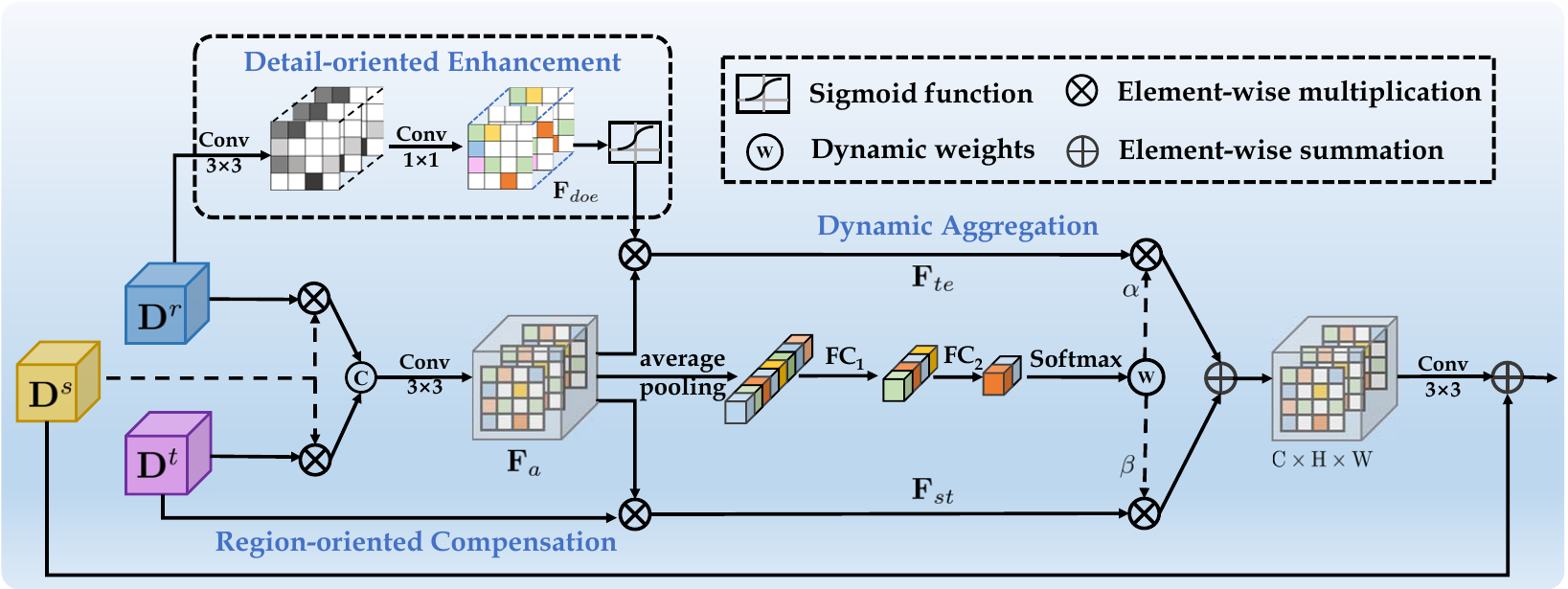} \\
	\end{tabular}
	\vspace{-1mm}
	\caption{Illustration of the Modality-aware Dynamic Aggregation Module (MDAM).}%
	\label{fig:MDFM}
	\vspace{-2mm}
\end{figure}

\subsection{Modality-Aware Dynamic Aggregation Module}\label{MDRM}
% ------------------------------------------
As illustrated in Figure~\ref{fig:network}, we exploit modality-shared union encoders and modality-induced feature modulator (MFM) to extract three groups of multi-level features. Consequently, three distinct flows, each with its individual responsibilities, are introduced to progressively aggregate features from top to bottom in order for the purpose of recovering complete salient objects. Given the potential impact of challenging environments on the quality of RGB and thermal inputs in real-world scenes, it is crucial to develop a robust RGB-T SOD model that can effectively mitigate interference from inferior modalities. However, directly and equally combining the information from diverse modalities may lead to uncontrollable and inconsistent results. To address this challenge, a Modality-aware Dynamic Aggregation Module~(MDAM) is introduced in our \textsc{ConTriNet} to integrate informative cues excavated from RGB- and thermal-modality flows into the modality-complementary flow in a dynamic weighting manner.

As shown in Figure~\ref{fig:MDFM}, given the acquired multi-level feature sets $\{D^{r}, D^{t}, D^{s}\}$, we initially perform an initial feature interaction and fusion by means of multiplication $\otimes$ and concatenation operations in order to obtain the aggregated features:
\begin{equation}	
	\begin{split}
		{F}_{a} = \textit{Conv}_{3\times 3}([{D}^{r}\otimes {D}^{s} ; {D}^{t}\otimes {D}^{s}]).
	\end{split}
\end{equation}
Empirically, RGB images harbor abundant color and texture details, while thermal images can highlight the contours and fine-grained texture of blurred regions in low-quality RGB images. Hence, we devise a detail-oriented enhancement branch and apply it to ${F}_{a}$ to guide the recovery of clear holistic topology ${F}_{te}$ by emphasizing areas with significant textural and structural variations. This detail-oriented enhancement process is defined as ${F}_{doe} = \mathcal{S}(\textit{Conv}_{1\times 1}(\textit{Conv}_{3\times 3}({D}^{r})))$, where $\mathcal{S}(\cdot)$ denotes the \textit{Sigmoid} activation function. For ${D}^{t}$, which encompasses additional semantic context cues, it is directly integrated with ${F}_{a}$ to incorporate region-level semantic compensation ${F}_{st}$. Here, ${F}_{te}={F}_{a}\otimes{F}_{doe}$ and ${F}_{st}={F}_{a}\otimes{D}^{t}$, which are referred to as modality-aware salient features. Subsequently, we perform dynamic aggregation by adaptively allocating the weights of the two features to fully exploit the intrinsic contextual relativity between paired modality-specific flows for bias reduction. This process is accomplished as follows:
\begin{equation}	
	\begin{split}
		{F}_{MDAM}^{s} = \textit{Conv}_{3\times 3}(\alpha \otimes {F}_{te}+\beta \otimes {F}_{st})+{D}^{s}.
	\end{split}
\end{equation}
The dynamic weights $\alpha$ and $\beta$ are calculated as follows:
\begin{equation}	
	\begin{split}
		\alpha, \beta = \textrm{Softmax}(\textrm{FC}_{2}(\textrm{FC}_{1}(\textrm{GAP}({F}_{a})))),
	\end{split}
\end{equation}
where $\textrm{GAP}(\cdot)$ denotes global average pooling. $\textrm{FC}_1(\cdot)$ and $\textrm{FC}_2(\cdot)$ represent two sequential fully connected layers that progressively refine the features for dynamic weighting, and $\textrm{Softmax}(\cdot)$ denotes the softmax function. The dynamic weights $\alpha$ and $\beta$ are determined based on the input cross-modality features. Furthermore, we compress the learning space by imposing a \textit{sum-to-one constraint} (\emph{i.e.,~}$\alpha+\beta=1$), which facilitates the learning of the MDAM. Overall, the process of MDAM is learnable and controls the contribution of complementary information of different modalities in accordance with the learned compensatory requirements, thereby facilitating the accurate recovery of complete salient objects even in highly complex scenes.

\subsection{Overall Loss Function}\label{sec:loss_function}
% ------------------------------------------
By constructing three sophisticated flows, we establish an end-to-end trainable confluent triple-flow network. This framework yields three predictions from the modality-specific RGB/Thermal flows and the modality-complementary flow, respectively. More specifically, the reconstructed features of the final layer in each flow are obtained through a $1\times 1$ convolution, upsampling operation, and sigmoid function, resulting in the individual saliency maps $\{M_{r}, M_{t}, M_{s} \}$. Here, $M_{r}$ and $M_{t}$ correspond to the modality-specific predictions, while $M_{s}$ represents the modality-complementary prediction. Moreover, by employing \textit{flow-cooperative fusion} (\emph{i.e.,} addition operation), we combine the aforementioned reconstructed predictions to generate a refined, comprehensive output denoted as $M_{f}$, which can be considered the final saliency map of our \textsc{ConTriNet}.
To optimize the proposed \textsc{ConTriNet} for modality-specific information preservation and modality-complementary information integration, the modality-specific flows and the modality-complementary flow are provided with a supervision signal. 
Taking into account the predicted saliency maps $\boldsymbol{M}=\{M_{r}, M_{t}, M_{s}, M_{f}\}$, inherited from~\cite{f3net_20}, we employ a combination of weighted binary cross-entropy (wBCE) loss $\mathcal{L}_{bce}^{w}$~\cite{BoerKMR05} and weighted IoU (wIoU) loss $\mathcal{L}_{iou}^{w}$~\cite{MattyusLU17} to impose constraints on them. The overall loss $L_{total}$ is formulated as:
\begin{equation} 
	\mathcal{L}_{total}=\sum_{{M}_{i} \in \boldsymbol{M}} \mathcal{L}_{bce}^{w}\left({M}_{i},G\right) + \sum_{{M}_{i} \in \boldsymbol{M}} \mathcal{L}_{iou}^{w}\left({M}_{i},{G}\right),
\end{equation}
where $G$ denotes the ground truth.

% ------------------------------------------
\begin{figure}[t!]
	\centering
	\begin{tabular}{c}
	\includegraphics[width=0.95 \linewidth]{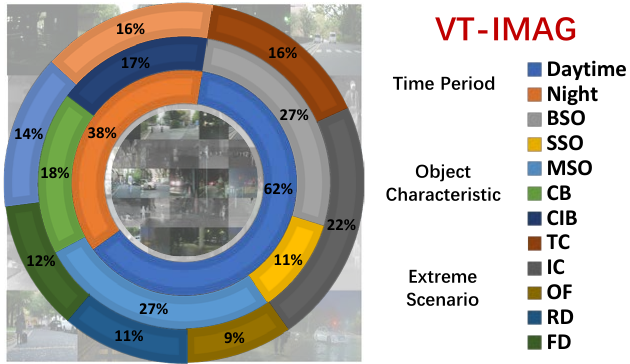} \\
	\end{tabular}
	\vspace{-2mm}
	\caption{The distribution of challenges in VT-IMAG, including time period (the inner circle), object characteristic (middle circle), and extreme scenario (outer circle).}%
	\label{fig:dataset}
 \vspace{-2mm}
\end{figure}
% ------------------------------------------
\section{Experiments}
\label{section4}

\begin{table*}[t!]
	\centering
	\caption{
		Comparison with recent state-of-the-art methods in RGB-D/RGB-T SOD on VT821~\cite{MTMR_18}, VT1000~\cite{SGDL_20} and VT5000~\cite{ADF_20}. The use of $\uparrow$/$\downarrow$ for a metric denotes that a larger/smaller value denotes better performance. The top three results are highlighted in a specific order: {\color{reda} \textbf{red}} for the best result, {\color{mygreen} \textbf{green}} for the second best, and {\color{myblue} \textbf{blue}} for the third best. \textsc{ConTriNet}$^{\tiny{16}}$ and \textsc{ConTriNet}$^{\tiny{50}}$ denote the utilization of VGG backbone~\cite{SimonyanZ14a} and ResNet backbone~\cite{GaoCZZYT21} as the encoder, respectively. \textsc{ConTriNet}$^{\star}$ represents the use of Swin-Transformer~\cite{LiuL00W0LG21} as the encoder.}
	\label{tab:cmp_1}
	\setlength\tabcolsep{0.6em}
	\resizebox{\linewidth}{!}{%
% 		\rowcolors{2}{gray!10}{white}
		\begin{tabular}{r*{3}{|*{5}{c}}}
	\toprule[2pt]
  & \multicolumn{5}{c|}{\textbf{VT821}} & \multicolumn{5}{c|}{\textbf{VT1000}} & \multicolumn{5}{c}{\textbf{VT5000}}                 
  \\ 
    \multirow{-2}{*}{\textbf{Method}}             
	& $S_{m}~\uparrow$ 
	& $F_{\beta}~\uparrow$  
	& $F^{\omega}_{\beta}~\uparrow$   
	& $E_{m}~\uparrow$ 
	& $\mathcal{M}~\downarrow$                             
	& $S_{m}~\uparrow$ 
	& $F_{\beta}~\uparrow$  
	& $F^{\omega}_{\beta}~\uparrow$   
	& $E_{m}~\uparrow$ 
	& $\mathcal{M}~\downarrow$                     
	& $S_{m}~\uparrow$ 
	& $F_{\beta}~\uparrow$  
	& $F^{\omega}_{\beta}~\uparrow$   
	& $E_{m}~\uparrow$ 
	& $\mathcal{M}~\downarrow$                  
	\\ 
	\midrule[1pt]
	\multicolumn{16}{c}{CNNs-Based Methods}
	\\
	\midrule[1pt]
	S2MA$_{20}$~\cite{S2MA_20}                       
	&0.829   &0.723    &0.702    &0.834    &0.081   
	&0.921   &0.852    &0.850    &0.914    &0.029                             
	&0.855   &0.751    &0.734    &0.869    &0.055 
	\\
	JL-DCF$_{20}$~\cite{JL-DCF_20}                       
	&0.839   &0.726    &0.720    &0.830    &0.076   
	&0.913   &0.829    &0.846    &0.899    &0.030                             
	&0.862   &0.738    &0.745    &0.860    &0.050 
	\\
	HAIN$_{21}$~\cite{HAINet_21}                       
	&0.849   &0.786    &0.758    &0.882    &0.046   
	&0.920   &0.878    &0.878    &0.927    &0.025                             
	&0.869   &0.808    &0.792    &0.900    &0.041 
	\\
	SPNet$_{21}$~\cite{SPNet_21}                       
	&0.864   &0.813    &0.809    &0.893    &0.038   
	&0.915   &0.880    &0.875    &0.939    &0.023                             
	&0.871   &0.834    &0.835    &0.902    &0.036 
	\\
	DCMF$_{22}$~\cite{DCMF_22}                       
	&0.856   &0.834    &0.740    &0.866    &0.055   
	&0.917   &0.834    &0.841    &0.915    &0.028            
	&0.857   &0.753    &0.728    &0.873    &0.052
        \\
        CIR-Net$_{22}$~\cite{CongLZLCHZ22}                       
	&0.861   &0.777    &0.724    &0.884    &0.045   
	&0.914   &0.861    &0.823    &0.924    &0.028                             
	&0.871   &0.791    &0.744    &0.899    &0.041
        % \\
 %        SPNet$_{23}$~\cite{ZhouFCZF23}                       
	% &{\color{mygreen} \textbf{0.891}}   &0.835    &{\color{myblue} \textbf{0.833}}    &0.912    &  {\color{mygreen} \textbf{0.031}}   
	% &{\color{mygreen} \textbf{0.929}}   &0.891    &0.894    &0.942    &{\color{myblue} \textbf{0.020}}  
	% &{\color{reda} \textbf{0.897}}   &{\color{myblue} \textbf{0.848}}    &{\color{mygreen} \textbf{0.838}}    &{\color{mygreen} \textbf{0.928}}    &{\color{reda} \textbf{0.030}}
        \\
        TBINet$_{22}$~\cite{WangZ22}                       
	&{\color{myblue} \textbf{0.890}}   &{\color{myblue} \textbf{0.842}}    &{\color{myblue} \textbf{0.834}}    &{\color{mygreen} \textbf{0.923}}    &{\color{mygreen} \textbf{0.030}}   
	&{\color{myblue} \textbf{0.931}}   &{\color{myblue} \textbf{0.899}}    &{\color{myblue} \textbf{0.899}}    &0.944    &{\color{myblue} \textbf{0.020}}                             
	&0.890   &0.844    &0.831    &{\color{mygreen} \textbf{0.925}}    &{\color{mygreen} \textbf{0.031}}
 %        \\
 %        SPSNet$_{22}$~\cite{LeePCL22}                       
	% &0.   &0.    &0.    &0.    &0.   
	% &0.   &0.    &0.    &0.    &0.                             
	% &0.   &0.    &0.    &0.    &0.
        \\
        RAFNet$_{22}$~\cite{WuGTAPD22}                       
	&0.883   &0.826    &0.811    &0.905    &0.038   
	&{\color{mygreen} \textbf{0.932}}   &0.893    &0.898    &0.941    &{\color{myblue} \textbf{0.020}}                             
	&{\color{myblue} \textbf{0.891}}   &0.840    &0.829    &0.921    &0.033
        \\
        PopNet$_{23}$~\cite{PopNet23}                       
	&0.861   &0.795    &0.774    &0.887    &0.043   
	&0.929   &0.893    &0.895    &0.939    &{\color{myblue} \textbf{0.020}}                             
	&0.887   &0.839    &0.827    &0.918    &0.034
        \\
        HiDAnet$_{23}$~\cite{WuAMMD23}                       
	&{\color{myblue} \textbf{0.890}}   &{\color{myblue} \textbf{0.842}}    &0.832    &{\color{myblue} \textbf{0.920}}    &{\color{myblue} \textbf{0.031}}   
	&0.927   &0.894    &0.897    &0.942    &{\color{myblue} \textbf{0.020}}                             
	&0.890   &{\color{myblue} \textbf{0.846}}    &{\color{mygreen} \textbf{0.836}}    &0.923    &{\color{myblue} \textbf{0.032}}
	\\
    \midrule
	MTMR$_{18}$~\cite{MTMR_18}                       
	&0.725   &0.662    &0.462    &0.815    &0.108   
	&0.706   &0.715    &0.485    &0.836    &0.119                             
	&0.680   &0.595    &0.387    &0.795    &0.114                         
	\\
	M3S-NIR$_{19}$~\cite{M3S-NIR_19}                       
	&0.723   &0.734    &0.407    &0.859    &0.140   
	&0.726   &0.717    &0.463    &0.827    &0.145                             
	&0.652   &0.575    &0.327    &0.780    &0.168
	\\
	SGDL$_{20}$~\cite{SGDL_20}                       
	&0.765   &0.730    &0.583    &0.847    &0.085   
	&0.787   &0.764    &0.652    &0.856    &0.090                             
	&0.750   &0.672    &0.558    &0.824    &0.089
	\\
	ADF$_{20}$~\cite{ADF_20}                       
	&0.810   &0.716    &0.626    &0.842    &0.077   
	&0.910   &0.847    &0.804    &0.921    &0.034                             
	&0.863   &0.778    &0.722    &0.891    &0.046
	\\
	MIDD$_{21}$~\cite{MIDD_21}                       
	&0.871   &0.804    &0.760    &0.895    &0.045   
	&0.907   &0.871    &0.848    &0.928    &0.029                             
	&0.856   &0.789    &0.753    &0.891    &0.046
	\\
	CSRNet$_{21}$~\cite{CSRN_21}                       
	&0.885  &0.830    &0.821    &0.908   &0.038   
	&0.918   &0.877    &0.878    &0.925    &0.024                             
	&0.868   &0.810    &0.796    &0.905    &0.042
	\\
	CGFNet$_{21}$~\cite{CGFNet_21}                       
	&0.879   &{\color{mygreen} \textbf{0.844}}    &0.826    &0.912    &0.036   
	&0.921   &{\color{mygreen} \textbf{0.903}}    &0.897    
   &0.941    &0.023       
	&0.882   &{\color{mygreen} \textbf{0.853}}    &0.831    &0.921    &0.035
	\\
	MMNet$_{21}$~\cite{MMNet_21}                       
	&0.873   &0.794    &0.783    &0.892    &0.040   
	&0.914   &0.861    &0.863    &0.923    &0.027                             
	&0.862   &0.780    &0.770    &0.887    &0.043
	% \\
 %        SPNet$_{21}$~\cite{SPNet_21}                       
	% &{\color{reda} \textbf{0.893}}   &0.835    &{\color{myblue} \textbf{0.833}}    &{\color{myblue} \textbf{0.912}}    &  {\color{mygreen} \textbf{0.031}}   
	% &{\color{mygreen} \textbf{0.929}}   &0.891    &0.894    &0.942    &{\color{myblue} \textbf{0.020}}  
	% &{\color{reda} \textbf{0.897}}   &{\color{myblue} \textbf{0.848}}    &{\color{mygreen} \textbf{0.838}}    &{\color{mygreen} \textbf{0.928}}    &{\color{reda} \textbf{0.030}}
        \\
	MIA$_{22}$~\cite{liang2022multi}                       
	&0.844   &0.740    &0.720    &0.850    &0.070   
	&0.924   &0.868    &0.864    &0.926    &0.025                             
	&0.878   &0.793    &0.780    &0.893    &0.040
	\\
	ECFFNet$_{22}$~\cite{ECFFNet_21}                       
	&0.877   &0.810    &0.801    &0.902    &0.034  
	&0.923   &0.876    &0.885    &0.930    &0.021                           
	&0.874   &0.806    &0.801    &0.906    &0.038
	\\
	OSRNet$_{22}$~\cite{OSRNet}                       
	&0.875   &0.813    &0.801    &0.896    &0.043   
	&0.926   &0.892    &0.891    &0.935    &0.022 
	&0.875   &0.823    &0.807    &0.908    &0.040
        \\
	LSNet$_{23}$~\cite{ZhouZLYY23}                       
	&0.878   &0.825    &0.809    &0.911    &0.033   
	&0.925   &0.885    &0.887    &0.935    &0.023 
	&0.877   &0.825    &0.806    &0.915    &0.037
 %        \\
	% WaveNet$_{23}$~\cite{ZhouSJCH23}                       
	% &0.   &0.    &0.    &0.    &0.   
	% &0.   &0.    &0.    &0.    &0. 
	% &0.   &0.    &0.    &0.    &0.
         \\
        CAVER$_{23}$~\cite{PangZZL23}                      
	&{\color{mygreen} \textbf{0.891}}   &0.839    &{\color{mygreen} \textbf{0.835}}    &0.919    &0.033
	&{\color{reda} \textbf{0.936}}   &{\color{mygreen} \textbf{0.903}}    &{\color{reda} \textbf{0.909}}    &{\color{myblue} \textbf{0.945}}    &{\color{reda} \textbf{0.017}}
	&{\color{mygreen} \textbf{0.892}}   &0.842
 &{\color{myblue} \textbf{0.835}}    &{\color{myblue} \textbf{0.924}}    &{\color{myblue} \textbf{0.032}}
	\\
    \midrule
    \rowcolor{gray!10} \textsc{ConTriNet}$^{\tiny{16}}$~(Ours)          
	&0.881  &0.831   &0.818   &0.912   &{\color{myblue} \textbf{0.031}} 
	&0.924  &0.897   &0.897   &{\color{mygreen} \textbf{0.946}}   &{\color{myblue} \textbf{0.020}} 
	&0.883  &0.840   &0.826   &0.921   &{\color{myblue} \textbf{0.032}}
	\\
	\rowcolor{gray!10} \textsc{ConTriNet}$^{\tiny{50}}$~(Ours) %        
	&{\color{reda} \textbf{0.892}}  &{\color{reda} \textbf{0.850}}   &{\color{reda} \textbf{0.840}}  &{\color{reda} \textbf{0.926}}   &{\color{reda} \textbf{0.027}}  
	&0.929  &{\color{reda} \textbf{0.906}}   &{\color{mygreen} \textbf{0.907}}   &{\color{reda} \textbf{0.948}}   &{\color{mygreen} \textbf{0.019}} 
	&{\color{reda} \textbf{0.894}}  &{\color{reda} \textbf{0.863}}   &{\color{reda} \textbf{0.848}}   &{\color{reda} \textbf{0.934}}   &{\color{reda} \textbf{0.030}}
	\\
	\midrule[1pt]
	\multicolumn{16}{c}{Transformer-Based Methods}
	\\
	\midrule[1pt]
	TriTransNet$_{21}$~\cite{TriTransNet_21}                       
	&0.896   &0.852    &{\color{mygreen} \textbf{0.867}}    &{\color{myblue} \textbf{0.927}}    &{\color{mygreen} \textbf{0.026}}   
	&0.933   &{\color{mygreen} \textbf{0.912}}   &{\color{myblue} \textbf{0.912}}    &{\color{mygreen} \textbf{0.947}}    &{\color{mygreen} \textbf{0.017}}
	&0.895   &0.851    &0.863    &0.932    &0.031
	\\
	SwinNet$_{22}$~\cite{SwinNet_22}                      
	&{\color{myblue} \textbf{0.904}}   &0.847    &0.818    &0.926    &0.030
	&{\color{myblue} \textbf{0.938}}   &0.896    &0.894    &{\color{mygreen} \textbf{0.947}}    &{\color{myblue} \textbf{0.018}}
	&{\color{mygreen} \textbf{0.912}}   &{\color{myblue} \textbf{0.865}}    &0.846    &{\color{myblue} \textbf{0.942}}    &{\color{myblue} \textbf{0.026}}
        \\
	HRTransNet$_{23}$~\cite{TangLTH23}                      
	&{\color{mygreen} \textbf{0.906}}   &{\color{myblue} \textbf{0.853}}    &0.849    &{\color{mygreen} \textbf{0.929}}    &{\color{mygreen} \textbf{0.026}}
	&{\color{myblue} \textbf{0.938}}   &0.900    &{\color{mygreen} \textbf{0.913}}    &{\color{myblue} \textbf{0.945}}    &{\color{mygreen} \textbf{0.017}}
	&{\color{mygreen} \textbf{0.912}}   &{\color{mygreen} \textbf{0.871}}    &{\color{mygreen} \textbf{0.870}}    &{\color{mygreen} \textbf{0.945}}    &{\color{mygreen} \textbf{0.025}}
        \\
	XMSNet$_{23}$~\cite{XMSNet23}                       
	&{\color{mygreen} \textbf{0.906}}   &{\color{mygreen} \textbf{0.859}}    &{\color{myblue} \textbf{0.859}}    &{\color{mygreen} \textbf{0.929}}    &{\color{myblue} \textbf{0.028}}   
	&{\color{mygreen} \textbf{0.936}}   &{\color{myblue} \textbf{0.903}}    &0.911    &{\color{myblue} \textbf{0.945}}    &{\color{myblue} \textbf{0.018}} 
	&{\color{myblue} \textbf{0.907}}   &{\color{mygreen} \textbf{0.871}}    &{\color{myblue} \textbf{0.865}}    &0.939    &0.028
	\\
	\midrule
    \rowcolor{gray!10} \textsc{ConTriNet}$^{\star}$~(Ours) % 	& {\color{myblue} \textbf{0.901}}         
	&{\color{reda} \textbf{0.915}}   &{\color{reda} \textbf{0.878}}    &{\color{reda} \textbf{0.876}}    &{\color{reda} \textbf{0.940}}    &{\color{reda} \textbf{0.022}}   
	&{\color{reda} \textbf{0.941}}   &{\color{reda} \textbf{0.918}}    &{\color{reda} \textbf{0.924}}    &{\color{reda} \textbf{0.954}}    &{\color{reda} \textbf{0.015}} 
	&{\color{reda} \textbf{0.923}}   &{\color{reda} \textbf{0.898}}    &{\color{reda} \textbf{0.895}}    &{\color{reda} \textbf{0.956}}   &{\color{reda} \textbf{0.020}}
	\\
	\bottomrule[2pt]
\end{tabular}
	}\vspace{-2mm}
\end{table*}

\begin{table}[!t]
	\centering
	\caption{
		Comparison of the complexity of some recent publicly available state-of-the-art RGB-T SOD methods.
	}
	\label{tab:cmp_more}
	\resizebox{\linewidth}{!}{%
		\begin{tabular}{r|r|c|c}
	\toprule[2pt]
	\textbf{Method} & \textbf{Backbone} & \textbf{FLOPs (G)} & \textbf{Params. (M)}
	\\ 
	\midrule[1pt]
	ADF$_{20}$~\cite{ADF_20}               
	& VGG-16~\cite{SimonyanZ14a}                        
	& 128.17         
	&  83.13              
	\\
	MIDD$_{21}$~\cite{MIDD_21}               
	& VGG-16~\cite{SimonyanZ14a}                        
	& 145.08         
	& 52.43               
	\\
	CGFNet$_{21}$~\cite{CGFNet_21}           
	& VGG-16~\cite{SimonyanZ14a}  
	& 231.08         
	& 66.38               
	\\ 
	OSRNet$_{22}$~\cite{OSRNet}           
	& VGG-16~\cite{SimonyanZ14a}   
	& 34.32         
	& 15.64               
	\\ 
	MMNet$_{21}$~\cite{MMNet_21}           
	& Res2Net-50~\cite{GaoCZZYT21}   
	& 53.72         
	& 64.12  
        \\ 
	% SPNet$_{21}$~\cite{SPNet_21}           
	% & Res2Net-50~\cite{GaoCZZYT21}   
	% & 135.86         
	% & 175.29  
 %        \\ 
	% CIR-Net$_{22}$~\cite{CongLZLCHZ22}           
	% & ResNet-50~\cite{HeZRS16}   
	% & 44.65         
	% & 103.48  
	% \\ 
        CAVER$_{23}$~\cite{PangZZL23}           
	& ResNet-50d~\cite{HeZ0ZXL19}   
	& 44.44         
	& 55.79  
	\\ 
	TriTransNet$_{21}$~\cite{TriTransNet_21}
	& ResNet50~\cite{HeZRS16}+ViT~\cite{DosovitskiyB0WZ21} 
	% & ResNet-50~\cite{GaoCZZYT21}+ViT-B~\cite{DosovitskiyB0WZ21}   
	& 292.34         
	& 138.72              
	\\ 
	SwinNet$_{22}$~\cite{SwinNet_22}
	&Swin-B~\cite{LiuL00W0LG21}   
	& 124.30      
	&   198.67 
        \\
        HRTransNet$_{23}$~\cite{TangLTH23}
        & ResNet18~\cite{HeZRS16}+HRFormer~\cite{YuanFHLZCW21}
	& 17.27      
	&  58.89
	\\ 
	\midrule[0.5pt]
	\rowcolor{gray!10} \textsc{ConTriNet}~(Ours)                   
	& VGG-16~\cite{SimonyanZ14a}               
	& 106.60          
	& 23.95    
	\\
	\rowcolor{gray!10} \textsc{ConTriNet}~(Ours)                                   
	& Res2Net-50~\cite{GaoCZZYT21}               
	& 37.12          
	& 34.77               
	\\
	\rowcolor{gray!10} \textsc{ConTriNet}~(Ours)                                  
	& Swin-B~\cite{LiuL00W0LG21}              
	& 126.88          
	& 96.31             
	\\
	\bottomrule[2pt]
\end{tabular}
	}\vspace{-2mm}
\end{table}

\subsection{Experimental setup}
% ----------------------------------------
\subsubsection{Datasets}
We perform extensive experimentation on three widely-used RGB-T benchmarks as well as our recently introduced challenging benchmark to comprehensively demonstrate the robustness of our \textsc{ConTriNet} under various challenging scenarios.

% ----------------------------------------
\myparagraph{Public datasets.} There exist three publicly accessible benchmark datasets for RGB-T SOD tasks: \textbf{VT821}~\cite{MTMR_18}, which comprises $821$ manually registered image pairs; \textbf{VT1000}~\cite{SGDL_20}, which consists of $1,000$ pairs of relatively simplistic scenes captured by highly aligned RGB and Thermal cameras; and \textbf{VT5000}~\cite{ADF_20}, which offers a collection of $5,000$ pairs of high-resolution, diverse, and minimally deviated images. To ensure fair comparisons, we follow the same training protocol as \cite{MIDD_21}, which involves utilizing the same $2,500$ image pairs from VT5000 for training purposes, while the remaining images, in conjunction with VT821 and VT1000, are employed for testing.

% ----------------------------------------
\myparagraph{The proposed VT-IMAG.} In order to enhance the robustness and broaden the applicability of existing RGB-T SOD algorithms in diverse real-world scenarios, we introduce a more challenging dataset, called \textbf{VT-IMAG}. This dataset comprises $536$ RGB images along with their corresponding thermal maps. To ensure wide diversity and comprehensive coverage, instead of capturing RGB-T images ourselves, we curate well-aligned RGB-T pairs from selected examples in an RGB-T semantic segmentation dataset~\cite{SunZL19} and an RGB-T object detection dataset~\cite{LiuFHWLZL22} commonly employed in the field of autonomous driving.
The image selection process involved four viewers independently identifying the most salient objects based on initial impressions, with a final selection achieved through consensus. Each chosen RGB-T image pair encompasses at least one salient object significant in both RGB and Thermal modalities. Professional annotators meticulously marked the ground truth masks for SOD on a pixel-by-pixel basis.

The \textbf{primary purpose} of the constructed VT-IMAG is to drive the advancement of RGB-T SOD methods and facilitate their deployment in real-world scenarios. This dataset overcomes the limitations of existing datasets, namely VT821, VT1000, and VT5000, which consist mostly of simple samples and common environments that are not suitable for analyzing the robustness of deep learning models. To address this, VT-IMAG is proposed with high-quality annotation, covering diverse object types and various scenarios commonly encountered in surveillance and autonomous driving (\eg,~vehicles, pedestrians, and roadblocks). 
As illustrated in the inner circle of Figure~\ref{fig:dataset}, the dataset structure is delineated, categorizing the $536$ aligned pairs in VT-IMAG into two types based on the time of day: Daytime and Night.
To encourage genuine object detection learning rather than mere memorization of object positions, a broad distribution of salient object locations and sizes is provided. As depicted in the middle circle of Figure~\ref{fig:dataset}, the dataset annotates a total of $599$ objects across five classes: big salient object (BSO), small salient object (SSO), multiple salient objects (MSO), center bias (CB), and cross image boundary (CIB). 
Additionally, as shown in the outer circle of Figure~\ref{fig:dataset}, considering the image acquisition process and distortions present in natural environments, all $536$ aligned pairs are annotated under seven challenging sub-scenarios: thermal crossover (TC), image clutter (IC), out of focus (OF), rainy day (RD), foggy day (FD), strong noise (SN), and similar appearance (SA). Figure~\ref{fig:dataset} provides a summary of the attribute distributions in the proposed VT-IMAG dataset.

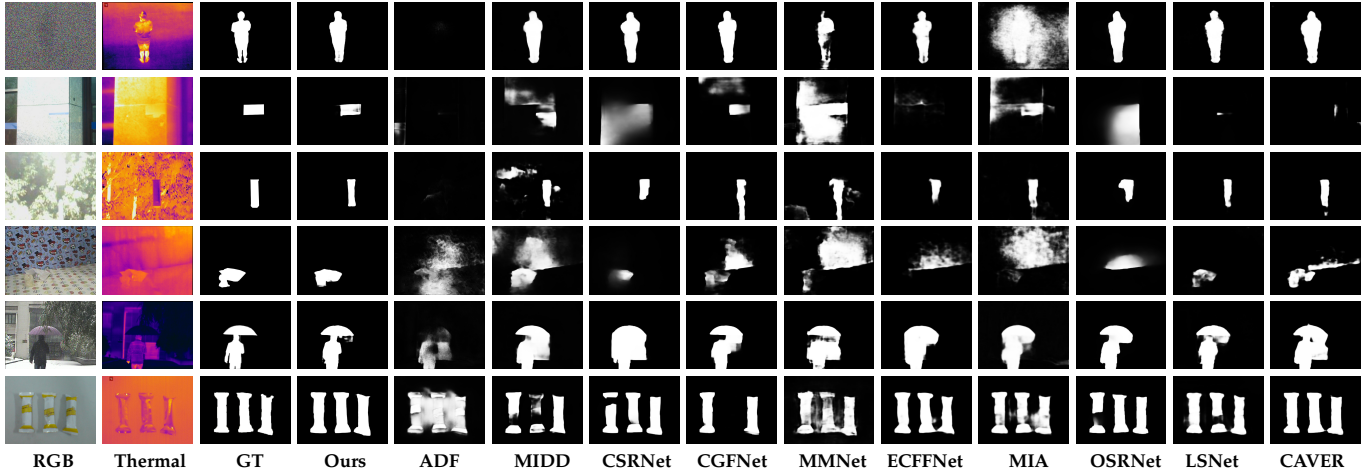
\begin{figure*}[tb!]
	\centering
	\begin{overpic}
        [width=0.99 \linewidth]{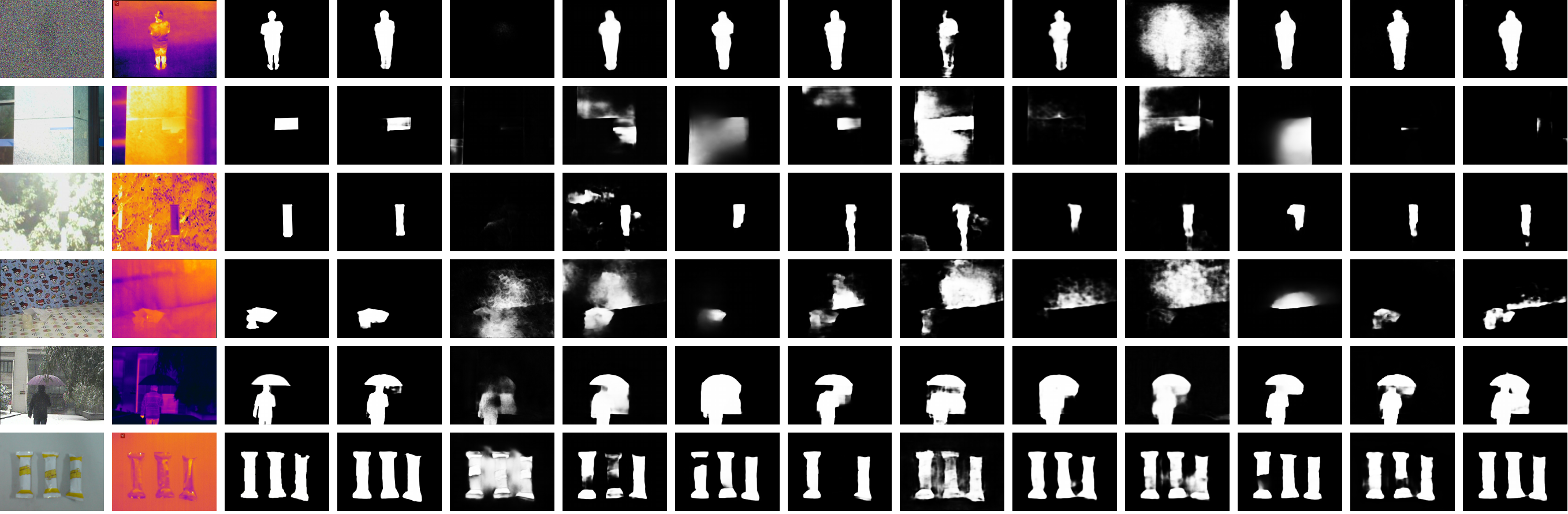}
        \put(2.0, -1.5){{\scriptsize \textbf{RGB}}}
        \put(8.0, -1.5){{\scriptsize \textbf{Thermal}}}
        \put(17.0, -1.5){{\scriptsize \textbf{GT}}}
        \put(23.5, -1.5){{\scriptsize \textbf{Ours}}}
        \put(30.5, -1.5){{\scriptsize \textbf{ADF}}}
        \put(37.5, -1.5){{\scriptsize \textbf{MIDD}}}
        \put(44, -1.5){{\scriptsize \textbf{CSRNet}}}
        \put(51, -1.5){{\scriptsize \textbf{CGFNet}}}
        \put(58.5, -1.5){{\scriptsize \textbf{MMNet}}}
        \put(65, -1.5){{\scriptsize \textbf{ECFFNet}}}
        \put(74, -1.5){{\scriptsize \textbf{MIA}}}
        \put(80, -1.5){{\scriptsize \textbf{OSRNet}}}
        \put(87, -1.5){{\scriptsize \textbf{LSNet}}}
        \put(94, -1.5){{\scriptsize \textbf{CAVER}}}
        \end{overpic}
        \vspace{2mm}
	\caption{Qualitative comparisons of visualization results are shown in the following columns: RGB image, Thermal image, Ground-Truth, and the predictions of ten state-of-the-art RGB-T SOD methods, respectively (zoomed-in for a better comparison).}
	\label{fig:examples}
	\vspace{-2mm}
\end{figure*}

\subsubsection{Evaluation Metrics}
In this work, we select five commonly-used SOD metrics for evaluation, including \textbf{S-measure} ($S_{m}$)~\cite{Sm}, \textbf{F-measure} ($F_{\beta}$)~\cite{Fm}, \textbf{Weighted F-measure} ($F_{\beta}^{w}$)~\cite{wFm}, \textbf{E-measure} ($E_{m}$)~\cite{Em}, and \textbf{MAE} ($\mathcal{M}$)~\cite{mae}. 
It is important to note that we report the mean F-measure and mean E-measure during our quantitative evaluation.
Higher values of $S_{m}, F_{\beta}, F_{\beta}^{w}, E_{m}$, and lower values of $\mathcal{M}$ indicate better performance.

\subsubsection{Implementation Details} 
The proposed method is implemented using the PyTorch platform, making use of a single NVIDIA GeForce RTX 3090 GPU. For training, the backbone networks are initialized with a pre-trained model on ImageNet~\cite{RussakovskyDSKS15}, while the remaining parameters of other modules are set to the PyTorch default values. The input RGB and Thermal images are resized to $352 \times 352$, and multiple augmentation techniques, such as random flipping, rotating, and border clipping, are employed to prevent overfitting. The network is trained using the Adam optimizer with a batch size of $16$, an initial learning rate of $5e-5$, and a cosine learning rate schedule. The final model converges within $100$ epochs. For testing, the predicted saliency maps are resized to the original size, and the fusion of three parallel saliency predictions is considered as the final prediction to achieve high-quality SOD performance.

\subsection{Comparisons with State-of-the-art Methods}
We compare the proposed \textsc{ConTriNet} with $27$ state-of-the-art methods, including ten \textit{CNNs-based RGB-D SOD} methods~(\emph{i.e.,} S2MA~\cite{S2MA_20}, JL-DCF~\cite{JL-DCF_20}, HAIN~\cite{HAINet_21}, SPNet~\cite{SPNet_21}, DCMF~\cite{DCMF_22}, CIR-Net~\cite{CongLZLCHZ22}, TBINet~\cite{WangZ22}, RAFNet~\cite{WuGTAPD22}, PopNet~\cite{PopNet23}, HiDAnet~\cite{WuAMMD23}),
three \textit{traditional RGB-T SOD} methods~(\emph{i.e.,} MTMR~\cite{MTMR_18}, M3S-NIR~\cite{M3S-NIR_19}, SDGL~\cite{SGDL_20}), and ten \textit{CNNs-based RGB-T SOD} methods~(\emph{i.e.,} ADF~\cite{ADF_20}, MIDD~\cite{MIDD_21}, CSRNet~\cite{CSRN_21}, CGFNet~\cite{CGFNet_21}, MMNet~\cite{MMNet_21}, ECFFN~\cite{ECFFNet_21}, MIA~\cite{liang2022multi}, OSRNet~\cite{OSRNet}, LSNet~\cite{ZhouZLYY23}, CAVER~\cite{PangZZL23}). 
We evaluate the proposed \textsc{ConTriNet} using both VGG16~\cite{SimonyanZ14a} and Res2Net50~\cite{GaoCZZYT21} backbones, denoted as \textsc{ConTriNet}$^{\tiny{16}}$ and \textsc{ConTriNet}$^{\tiny{50}}$. To overcome the performance bottleneck of RGB-T SOD, we introduce \textsc{ConTriNet}$^{\star}$, which adopts a stronger backbone, Swin Transformer~\cite{LiuL00W0LG21}, as the encoder. We compare \textsc{ConTriNet}$^{\star}$ with the recent \textit{Transformer-based multi-modality} methods, \ie, TriTransNet~\cite{TriTransNet_21} SwinNet~\cite{SwinNet_22}, HRTransNet~\cite{TangLTH23}, and XMSNet~\cite{XMSNet23}. For fair comparisons, RGB-D SOD methods are retrained on RGB-T datasets with their default settings, and the saliency maps provided by the RGB-T SOD methods are directly used for comparison.

\begin{table}[tb!]
	\centering
	\caption{
		Comparison with recent publicly available state-of-the-art RGB-T SOD methods on the proposed VT-IMAG. $\uparrow$/$\downarrow$ for a metric denotes that a larger/smaller value is better.
	}\vspace{-1mm}
	\label{tab:cmp_IMAG}
	\resizebox{\linewidth}{!}{%
		\begin{tabular}{r*{1}{|*{5}{c}}}
	\toprule[2pt]
  & \multicolumn{5}{c}{\textbf{VT-IMAG}}          
  \\ 
    \multirow{-2}{*}{\textbf{METHOD}}             
	& $S_{m}~\uparrow$ 
	& $F_{\beta}~\uparrow$  
	& $F^{\omega}_{\beta}~\uparrow$   
	& $E_{m}~\uparrow$ 
	& $\mathcal{M}~\downarrow$                             
	\\ 
	\midrule[1pt]
	ADF$_{20}$~\cite{ADF_20}                       
	&0.693   &0.506    &0.404   &0.729    &0.105   
	\\
	MIDD$_{21}$~\cite{MIDD_21}                       
	&0.761   &0.609    &0.534    &0.797    &0.063   
	\\
	CSRNet$_{21}$~\cite{CSRN_21}                       
	&0.601  &0.388    &0.314   &0.713  &0.097
	\\
	CGFNet$_{21}$~\cite{CGFNet_21}                       
	&0.795   &0.696    &0.664    &0.861     &0.042   
	\\
	OSRNet$_{22}$~\cite{OSRNet}                       
	&0.782   &0.665    &0.626   &0.839    &0.051  
        \\
	LSNet$_{23}$~\cite{ZhouZLYY23}                       
	&0.779   &0.712    &0.636   &0.896    &0.038
        \\
	CAVER$_{23}$~\cite{PangZZL23}                       
	&0.815   &0.724    &0.693   &0.885    &0.032 
	\\
        \midrule
	TriTransNet$_{21}$~\cite{TriTransNet_21}     
	&0.805   &0.704    &0.681    &0.866    &0.037   
	\\
	SwinNet$_{22}$~\cite{SwinNet_22}                      
	&0.838   &0.757    &0.669   &0.910    &0.032
	\\
        HRTransNet$_{23}$~\cite{TangLTH23}
        &0.848   &0.769    &0.738   &0.912    &0.027
        \\
        XMSNet$_{23}$~\cite{XMSNet23}
        &0.833   &0.744    &0.722   &0.903    &0.029
        \\
    \midrule
    \rowcolor{gray!10} \textsc{ConTriNet}$^{\tiny{16}}$~(Ours)          
	&0.801  & 0.729  &0.682   & 0.886 &  0.032
	\\
	\rowcolor{gray!10} \textsc{ConTriNet}$^{\tiny{50}}$~(Ours) %        
	&0.828 &0.745   &0.727 &0.902   & 0.029
	\\
    \rowcolor{gray!10} \textsc{ConTriNet}$^{\star}$~(Ours) %          
	&\textbf{0.868}   &\textbf{0.832}   &\textbf{0.804}    &\textbf{0.943}    & \textbf{0.021}
	\\
	\bottomrule[2pt]
\end{tabular}
	}\vspace{-2mm}
\end{table}

\newcommand{\AddImg}[1]{%
    \includegraphics[width=.075\textwidth]{VT-IMAG/#1_RGB.png} &%
    \includegraphics[width=.075\textwidth]{VT-IMAG/#1_T.png} &%
    \includegraphics[width=.075\textwidth]{VT-IMAG/#1_GT.png} &%
    \includegraphics[width=.075\textwidth]{VT-IMAG/#1_Ours.png} &%
    \includegraphics[width=.075\textwidth]{VT-IMAG/#1_MIDD.png} &%
    \includegraphics[width=.075\textwidth]{VT-IMAG/#1_CSRNet.png} &%
    \includegraphics[width=.075\textwidth]{VT-IMAG/#1_CGFNet.png} &%
    \includegraphics[width=.075\textwidth]{VT-IMAG/#1_OSRNet.png} &%
    \includegraphics[width=.075\textwidth]{VT-IMAG/#1_LSNet.png} &%
    \includegraphics[width=.075\textwidth]{VT-IMAG/#1_CAVER.png} &%
    \includegraphics[width=.075\textwidth]{VT-IMAG/#1_TriTransNet.png} &%
    \includegraphics[width=.075\textwidth]{VT-IMAG/#1_SwinNet.png}& %
    \includegraphics[width=.075\textwidth]{VT-IMAG/#1_HRTransNet.png} &
    \includegraphics[width=.075\textwidth]{VT-IMAG/#1_XMSNet.png} %
}

\begin{figure*}[!th]
    \centering
    \footnotesize
    \resizebox{\textwidth}{!} {
    % \begin{tabular}{ccccccccccccc}
    \begin{tabular}{m{1.0cm}<{\centering}m{1.00cm}<{\centering}m{1.00cm}<{\centering}m{1.00cm}<{\centering}m{1.00cm}<{\centering}m{1.00cm}<{\centering}m{1.00cm}<{\centering}m{1.00cm}<{\centering}m{1.00cm}<{\centering}m{1.00cm}<{\centering}m{1.00cm}<{\centering}m{1.00cm}<{\centering}m{1.00cm}<{\centering}m{1.00cm}<{\centering}m{1.00cm}<{\centering}}
    \textbf{\tiny Daytime}&\AddImg{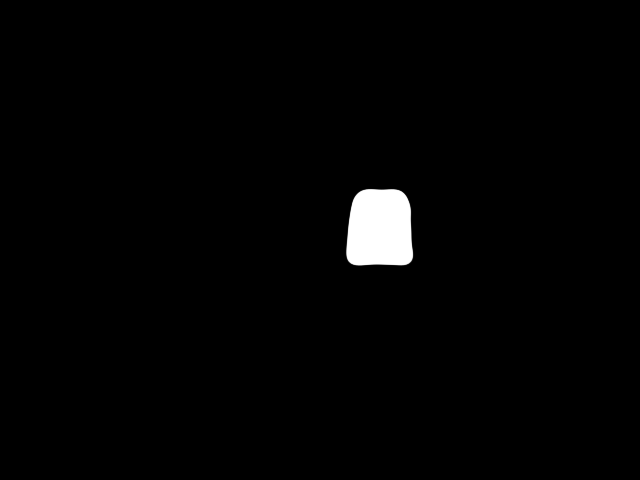}\\
    \textbf{\tiny Night} & \AddImg{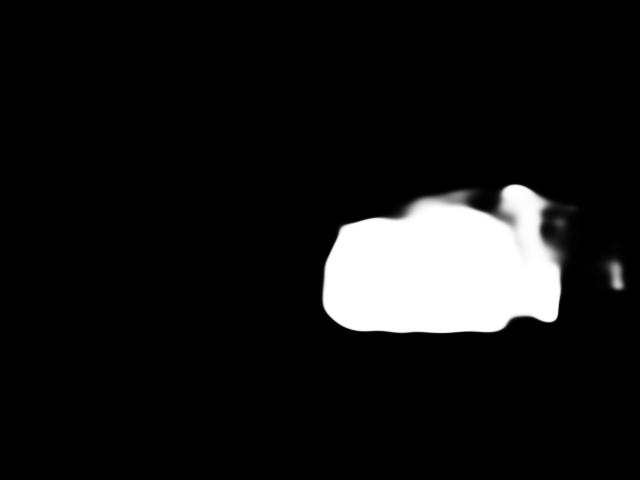} \\
    \textbf{\tiny Big Salient Object} & \AddImg{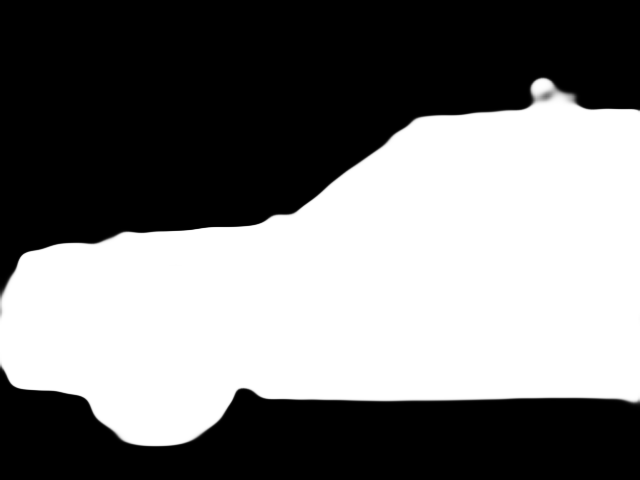} \\
    \textbf{\tiny Small Salient Object} & \AddImg{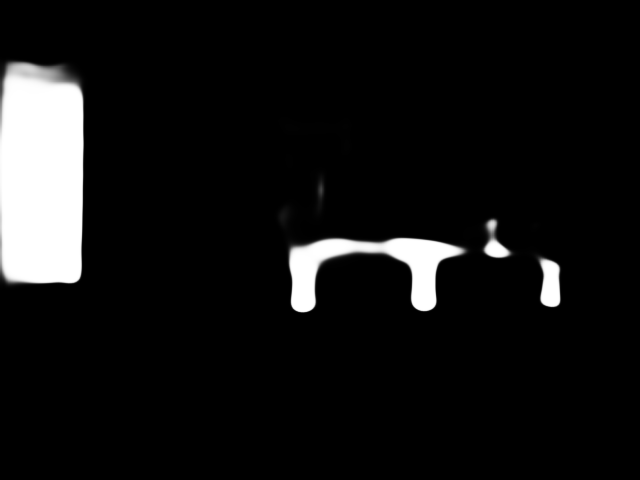} \\
    \textbf{\tiny Multiple Salient Object} & \AddImg{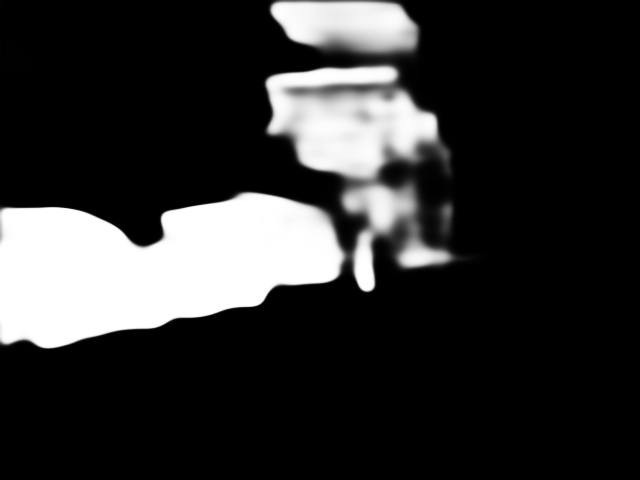} \\
    \textbf{\tiny Center Bias} & \AddImg{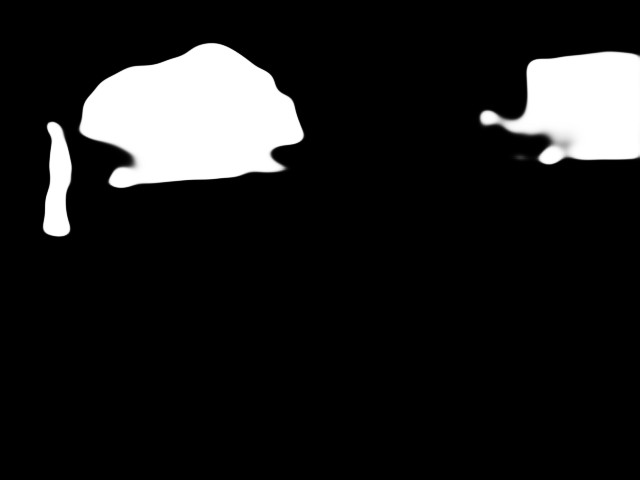} \\
    \textbf{\tiny Cross Image Boundary} & \AddImg{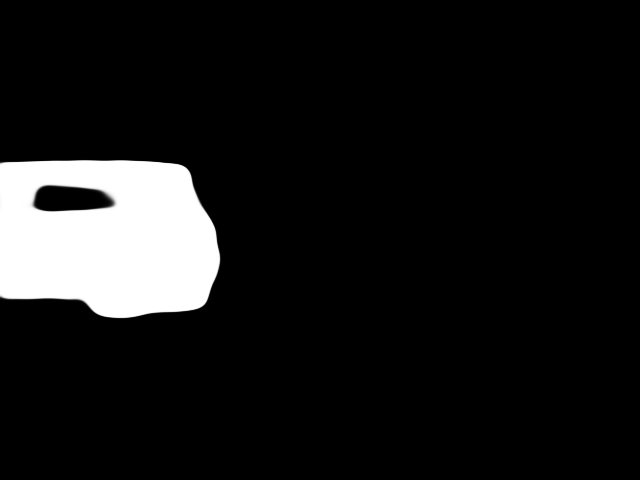} \\
        \textbf{\tiny Thermal Crossover} & \AddImg{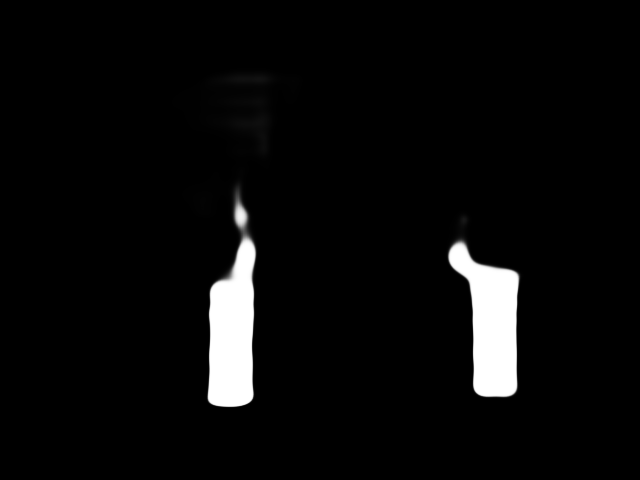} \\
        \textbf{\tiny Image Clutter} &\AddImg{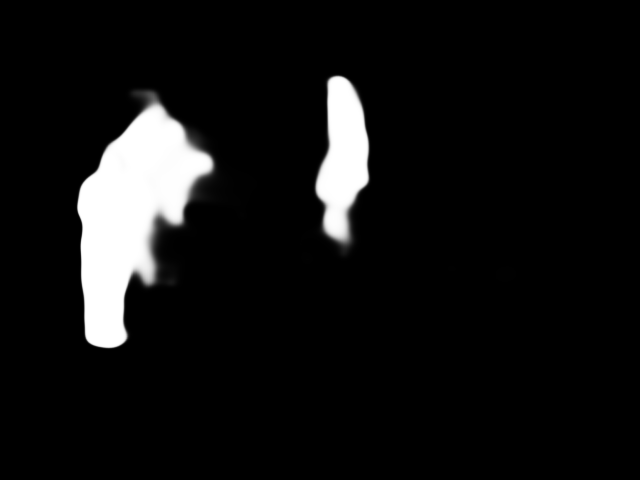} \\
         \textbf{\tiny Out of Focus} &\AddImg{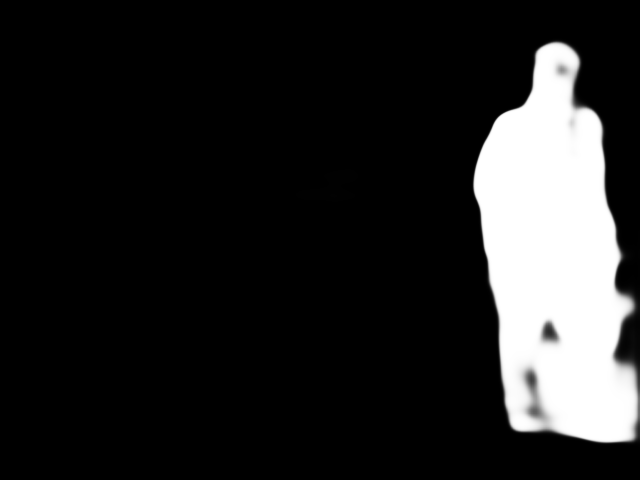} \\
         \textbf{\tiny Rainy Day} &\AddImg{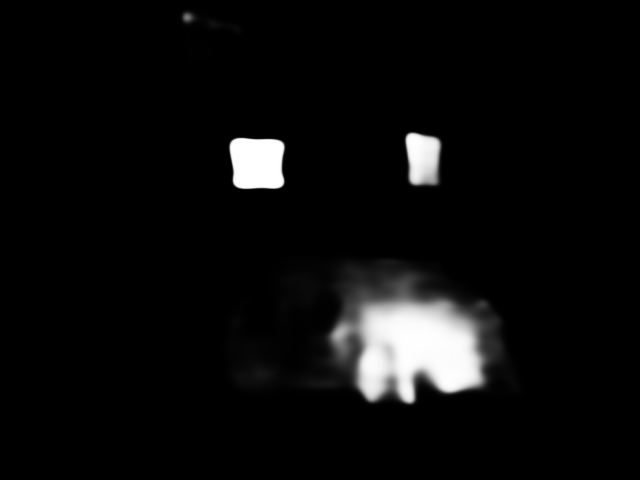} \\
         \textbf{\tiny Foggy Day} &\AddImg{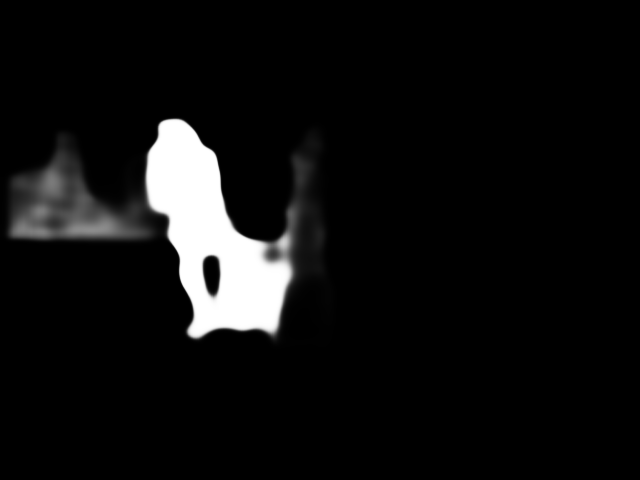} \\
        \textbf{\tiny Strong Noise}  &\AddImg{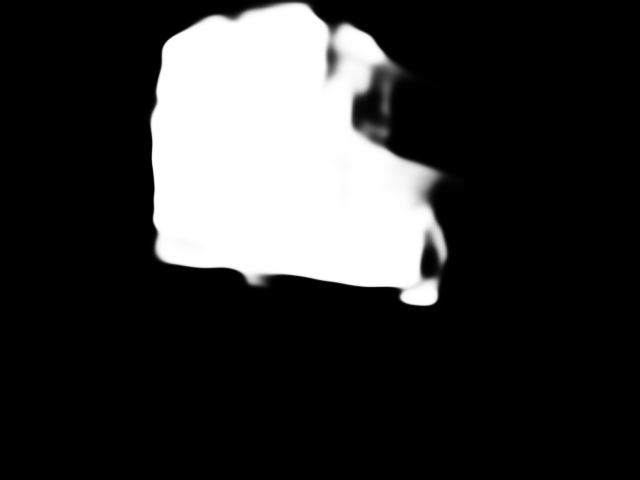} \\
         \textbf{\tiny Similar Appearance} &\AddImg{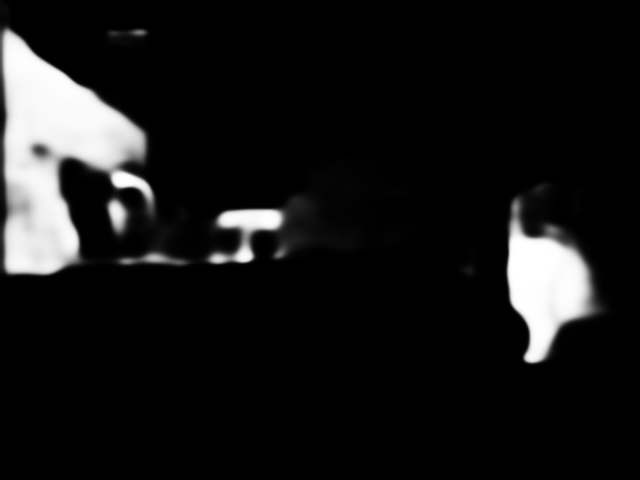} \\
       ~ & {\scriptsize \textbf{RGB}} & {\scriptsize \textbf{Thermal}} & {\scriptsize \textbf{GT}} & {\scriptsize \textbf{Ours}} & {\scriptsize \textbf{MIDD}} & {\scriptsize  \textbf{CSRNet}} & {\scriptsize \textbf{CGFNet}} & {\scriptsize \textbf{OSRNet}} & {\scriptsize \textbf{LSNet}} & {\scriptsize \textbf{CAVER}}  & {\scriptsize \textbf{TriTransNet}} & {\scriptsize \textbf{SwinNet}} & {\scriptsize \textbf{HRTransNet}} & {\scriptsize \textbf{XMSNet}}
    \end{tabular}}
    \vspace{-1mm}
    \caption{Qualitative comparisons of our proposed method and recent publicly available RGB-T SOD methods on various challenges in the proposed VT-IMAG (zoomed-in for a better comparison).}
    \label{fig:VT-IMAG}
\end{figure*}%

\begin{table*}[t!]
    \centering
    \caption{Ablation analysis of the different components with their various design choices on three datasets.}
    \vspace{-1mm}
    \label{tab:cmp_MFM}%
    \resizebox{\linewidth}{!}{%
    \begin{tabular}{c|l*{3}{|*{5}{c}}}
	\toprule[2pt]
    & & \multicolumn{5}{c|}{\textbf{VT821}} & \multicolumn{5}{c|}{\textbf{VT1000}} & \multicolumn{5}{c}{\textbf{VT5000}}                                                                                  \\
	\multirow{-2}{*}{\textbf{No.}}
	&\multirow{-2}{*}{\textbf{Settings}}             
	& $S_{m}~\uparrow$ 
	& $F_{\beta}~\uparrow$  
	& $F^{\omega}_{\beta}~\uparrow$   
	& $E_{m}~\uparrow$ 
	& $\mathcal{M}~\downarrow$                             
	& $S_{m}~\uparrow$ 
	& $F_{\beta}~\uparrow$  
	& $F^{\omega}_{\beta}~\uparrow$   
	& $E_{m}~\uparrow$ 
	& $\mathcal{M}~\downarrow$                     
	& $S_{m}~\uparrow$ 
	& $F_{\beta}~\uparrow$  
	& $F^{\omega}_{\beta}~\uparrow$   
	& $E_{m}~\uparrow$ 
	& $\mathcal{M}~\downarrow$                
	\\ 
	\midrule[1pt]
	1 & Baseline (w/~flow$\times$1)                                  
	& 0.864 & 0.799 & 0.781 & 0.896 & 0.038  
	& 0.919 & 0.884 & 0.884 & 0.933 & 0.023
	& 0.876 & 0.829 & 0.813 & 0.919 & 0.035
	\\
	\midrule
	2 & No.~1+MFM (w/o AFF)                                    
	& 0.872 & 0.813 & 0.803 & 0.909 & 0.035 
	& 0.921 & 0.891 & 0.895 & 0.940 & 0.021
	& 0.881 & 0.841 & 0.824 & 0.926 & 0.034 
% 	& 0.883 & 0.845 & 0.822 & 0.924 & 0.033 
	\\
	3 & No.~1+MFM (w/o CFE)                                    
	& 0.871 & 0.811 & 0.799 & 0.907 & 0.037 
	& 0.920 & 0.890 & 0.898 & 0.941 & 0.021
	& 0.883 & 0.845 & 0.822 & 0.924 & 0.033  
	\\
	4 & No.~1+MFM (default)                                    
	& 0.876 & 0.822 & 0.812 & 0.910 & 0.034
	& 0.924 & 0.895 & 0.900 & 0.945 & 0.021
	& 0.884 & 0.848 & 0.834 & 0.928 & 0.032
	\\
	\midrule
	5 & No.~4+flow$\times$2
	& 0.883 & 0.834 & 0.820 & 0.913 & 0.032
	& 0.926 & 0.899 & 0.901 & 0.943 & 0.020
	& 0.890 & 0.855 & 0.841 & 0.929 & 0.031
	\\
	6 & No.~5+RASPM 
	& 0.885 & 0.837 & 0.828 & 0.914 & 0.030
	& 0.927 & 0.902 & 0.902 & 0.946 & 0.020
	& 0.889 & 0.857 & 0.843 & 0.932 & {\textbf{0.030}}
	\\
	\midrule
	7 & No.~6+MDAM (w/o DW) 
	& 0.890 & 0.848 & 0.836 & 0.924 & 0.029
	& 0.927 & 0.905 & 0.905 & {\textbf{0.948}} & 0.020
	& 0.892 & 0.860 & 0.845 & {\textbf{0.934}} & {\textbf{0.030}}
	\\
        8 & No.~6+MDAM (w/o DoE) 
	& 0.886 & 0.847 & 0.834 & 0.921 & 0.030
	& 0.928 & 0.904 & 0.905 & {\textbf{0.948}} & {\textbf{0.019}}
	& 0.893 & 0.861 & 0.846 & {\textbf{0.934}} & {\textbf{0.030}}
	\\
	\rowcolor{gray!10} 9 & No.~6+MDAM (default) 
	&{\textbf{0.892}}   &{\textbf{0.850}}    &{\textbf{0.840}}    &{\textbf{0.926}}    &{\textbf{0.027}}   
	&{\textbf{0.929}}   &{\textbf{0.906}}    &{\textbf{0.907}}    &{\textbf{0.948}}    &{\textbf{0.019}} 
	&{\textbf{0.894}}   &{\textbf{0.863}}    &{\textbf{0.848}}    &{\textbf{0.934}}   &{\textbf{0.030}}
	\\
	\bottomrule[2pt]
\end{tabular}

      }\vspace{-1mm}
\end{table*}%

\subsubsection{Quantitative Comparison}
Table~\ref{tab:cmp_1} presents a summary of quantitative comparisons across three benchmarks utilizing five evaluation metrics. The proposed \textsc{ConTriNet} outperforms three traditional methods (\ie,~\cite{MTMR_18, M3S-NIR_19, SGDL_20}) by a large margin and demonstrates competitive performance compared to all state-of-the-art CNNs-based methods, achieving the best results on most evaluation indicators. Especially, \textsc{ConTriNet}$^{\tiny{50}}$ shows remarkable superiority on the \textbf{VT821} dataset, which contains more disturbing information than the other two datasets.
When compared with the second-best method, \textsc{ConTriNet}$^{\tiny{50}}$ achieves a minimum percentage gain of $1.2\%$ for $F_{\beta}$, $ 1.2\%$ for $F_{\beta}^{w}$, $1.1\%$ for $E_{m}$, and $10.0\%$ for $\mathcal{M}$ score. We also retrain seven state-of-the-art RGB-D SOD models and extend them to RGB-T SOD tasks. However, most RGB-D methods perform poorly and the dramatic drop in performance indicates that these two tasks are not fully compatible. 
Moreover, the stronger version of \textsc{ConTriNet}, \ie,~\textsc{ConTriNet}$^{\star}$, holds absolute advantages over recent Transformer-based methods, although the recent Transformer-based methods HRTransNet~\cite{TangLTH23} and XMSNet~\cite{XMSNet23} achieve superior performance compared to state-of-the-art CNNs-based methods. Furthermore, we compare the complexity of the proposed \textsc{ConTriNet} with recent publicly state-of-the-art RGB-T SOD methods, as shown in Table~\ref{tab:cmp_more}. Across different backbones, \textsc{ConTriNet} demonstrates a relatively small number of model parameters and competitive computational efficiency. Unlike existing methods that adopt modality-specific encoders for two-stream frameworks, the efficacy and efficiency of our \textsc{ConTriNet} stem from the proposed ``Divide-and-Conquer'' strategy, which deserves more attention in future work for robust RGB-T SOD with higher efficiency.

\begin{table*}[t!]
    \centering
    \caption{Comparison of the proposed RASPM with other alternatives on three datasets.}
    \label{tab:cmp_RASP}%
     \vspace{-1mm}
    \resizebox{\linewidth}{!}{%
    \begin{tabular}{c|c*{3}{|*{5}{c}}}
	\toprule[2pt]
  & & \multicolumn{5}{c|}{\textbf{VT821}} & \multicolumn{5}{c|}{\textbf{VT1000}} & \multicolumn{5}{c}{\textbf{VT5000}}                 
  \\ 
    \multirow{-2}{*}{\textbf{No.}}
    &\multirow{-2}{*}{\textbf{Settings}}             
	& $S_{m}~\uparrow$ 
	& $F_{\beta}~\uparrow$  
	& $F^{\omega}_{\beta}~\uparrow$   
	& $E_{m}~\uparrow$ 
	& $\mathcal{M}~\downarrow$                             
	& $S_{m}~\uparrow$ 
	& $F_{\beta}~\uparrow$  
	& $F^{\omega}_{\beta}~\uparrow$   
	& $E_{m}~\uparrow$ 
	& $\mathcal{M}~\downarrow$                     
	& $S_{m}~\uparrow$ 
	& $F_{\beta}~\uparrow$  
	& $F^{\omega}_{\beta}~\uparrow$   
	& $E_{m}~\uparrow$ 
	& $\mathcal{M}~\downarrow$                  
	\\
	\midrule[1pt]
        1& Conv ($3\times 3$)                     
	&0.883   &0.836    &0.821    &0.916    &0.032   
	&0.925   &0.899    &0.898    &0.945    &0.021     
	&0.886   &0.850    &0.835    &0.929    &0.032 
	\\
	2& PPM~\cite{PPM_17}                     
	&0.881   &0.838    &0.823    &0.914    &0.032   
	&0.929   &0.904    &0.905    &0.947    &{\textbf{0.019}}     
	&0.885   &0.849    &0.833    &0.927    &0.032 
	\\
	3& ASPP~\cite{ASPP_18}                      
	&0.886   &0.842    &0.830    &0.918    &0.030   
	&0.927   &0.905    &0.903    &0.947    &{\textbf{0.019}}      
	&0.889   &0.856    &0.840    &0.931    &0.031 
	\\
	4& RASPM (w/o atrous)
	&0.885   &0.842    &0.828    &0.915    &0.031 
	&{\textbf{0.929}}   &{\textbf{0.907}}    & {0.906}  &{\textbf{0.948}}    &{\textbf{0.019}} 
	&0.893   &0.859    &0.845    &0.932    &{\textbf{0.030}}
	\\
        \rowcolor{gray!10} 5& RASPM (default) % 	& {\color{myblue} \textbf{0.901}}         
	&{\textbf{0.892}}   &{\textbf{0.850}}    &{\textbf{0.840}}    &{\textbf{0.926}}    &{\textbf{0.027}}   
	&{\textbf{0.929}}   &{0.906}    &{\textbf{0.907}}    &{\textbf{0.948}}    &{\textbf{0.019}} 
	&{\textbf{0.894}}   &{\textbf{0.863}}    &{\textbf{0.848}}    &{\textbf{0.934}}   &{\textbf{0.030}}
	\\
	\bottomrule[2pt]
\end{tabular}
      }
      % \vspace{-1mm}
 \end{table*}%
\begin{table*}[t!]
    \centering
    \caption{Comparison of the proposed \textsc{ConTriNet} with different loss functions on three datasets.}
    \label{tab:cmp_loss}%
     \vspace{-1mm}
    \resizebox{\linewidth}{!}{%
    \begin{tabular}{c|c*{3}{|*{5}{c}}}
	\toprule[2pt]
    & & \multicolumn{5}{c|}{\textbf{VT821}} & \multicolumn{5}{c|}{\textbf{VT1000}} & \multicolumn{5}{c}{\textbf{VT5000}}                                                                                  \\
	\multirow{-2}{*}{\textbf{No.}}
	&\multirow{-2}{*}{\textbf{Loss Settings}}             
	& $S_{m}~\uparrow$ 
	& $F_{\beta}~\uparrow$  
	& $F^{\omega}_{\beta}~\uparrow$   
	& $E_{m}~\uparrow$ 
	& $\mathcal{M}~\downarrow$                             
	& $S_{m}~\uparrow$ 
	& $F_{\beta}~\uparrow$  
	& $F^{\omega}_{\beta}~\uparrow$   
	& $E_{m}~\uparrow$ 
	& $\mathcal{M}~\downarrow$                     
	& $S_{m}~\uparrow$ 
	& $F_{\beta}~\uparrow$  
	& $F^{\omega}_{\beta}~\uparrow$   
	& $E_{m}~\uparrow$ 
	& $\mathcal{M}~\downarrow$                
	\\ 
	\midrule[1pt]
	1 & wBCE~($\mathcal{L}_{bce}^{w}$) only                                  
	& {\textbf{0.894}} & 0.830 & 0.830 & 0.909 & 0.033  
	& {\textbf{0.934}} & 0.895 & 0.904 & 0.941 & {\textbf{0.018}}
	& {\textbf{0.895}} & 0.840 & 0.836 & 0.922 & {\textbf{0.030}}
	\\
	2 & wIoU~($\mathcal{L}_{iou}^{w}$) only                                    
	& 0.879 & 0.840 & 0.825 & 0.910 & 0.032  
	& 0.922 & 0.903 & 0.899 & 0.944 & 0.020
	& 0.887 & 0.857 & 0.842 & 0.932 & 0.031 
	\\
	\rowcolor{gray!10} 3 & wBCE+wIoU~($\mathcal{L}_{bce}^{w}$+$\mathcal{L}_{iou}^{w}$)    
	&0.892  &{\textbf{0.850}}    &{\textbf{0.840}}    &{\textbf{0.926}}    &{\textbf{0.027}}   
	&0.929   &{\textbf{0.906}}    &{\textbf{0.907}}    &{\textbf{0.948}}    &0.019 
	&0.894   &{\textbf{0.863}}    &{\textbf{0.848}}    &{\textbf{0.934}}   &{\textbf{0.030}} 
	\\
	\bottomrule[2pt]
\end{tabular}

      }
 % \vspace{-1mm}
\end{table*}%

\subsubsection{Qualitative Comparison}
Figure~\ref{fig:examples} presents saliency maps produced by recent representative RGB-T SOD methods, showcasing their performance in several challenging scenarios: strong noises ($1^{st}$ row), thermal crossover ($2^{nd}$ row), overexposure ($3^{rd}$ row), complex background ($4^{th}$ row), bad weather ($5^{th}$ row) and multiple objects ($6^{th}$ row). Existing methods fail to ensure structural integrity for low-quality and defective modality data, unlike our \textsc{ConTriNet}, which achieves this through powerful modality-specific mining and complementation offered by the triple-flow paradigm. 
Additionally, while competitors cannot locate the salient object in the complex backgrounds of the $4^{th}$ row, our model accurately identifies the conspicuous object with clear boundaries, facilitated by the dynamic exploitation of saliency-related cues provided by the proposed MDAM.
In multi-object scenarios, our model effectively reduces omission and prevents object adhesion, mainly due to the refinement of the predicted saliency object by our proposed RASPM, resulting in sharper boundaries. Furthermore, \textsc{ConTriNet} demonstrates robustness to bad weather conditions, as demonstrated in the $5^{th}$ row, thanks to the efficient refinement and integration of multi-modal information by the proposed MFM. Overall, \textsc{ConTriNet} produces more favorable salient maps with complete objects and precise boundaries, affirming its robustness and effectiveness across diverse scenes and objects.

\subsection{Robustness Analysis}
To demonstrate the superior robustness of the proposed \textsc{ConTriNet}, we evaluate the performances of different deep-learning models on the VT-IMAG dataset. Table~\ref{tab:cmp_IMAG} presents the quantitative comparison of recent state-of-the-art RGB-T SOD methods, including ADF~\cite{ADF_20}, MIDD~\cite{MIDD_21}, CSRNet~\cite{CSRN_21}, CGFNet~\cite{CGFNet_21}, OSRNet~\cite{OSRNet}, LSNet~\cite{ZhouZLYY23}, CAVER~\cite{PangZZL23}, TriTransNet~\cite{TriTransNet_21}, SwinNet~\cite{SwinNet_22}), HRTransNet~\cite{TangLTH23}, and XMSNet~\cite{XMSNet23}. Considering the disparities between training data and real-world scenarios, the sensitivity to unknown challenging cases better reflects the model's robustness. Hence, to ensure a fair comparison, all models are solely trained on clear data and simple scenes (\ie,~training set of VT5000) and evaluated for \textbf{zero-shot robustness} on various real-world challenging cases in VT-IMAG. 
As shown in Table~\ref{tab:cmp_IMAG}, the proposed \textsc{ConTriNet} significantly outperforms all competitors, exhibiting consistent and substantial performance gains across the five evaluation indicators. For example, ``\textsc{ConTriNet}$^{\star}$'' achieves the best performance (\ie,~$S_{m}: 0.868$, $F_{\beta}: 0.832$, $F_{\beta}^{w}: 0.804$, $E_{m}: 0.943$, and $\mathcal{M}: 0.021$) with significant relative gains of $1.9\%$, $8.2\%$, $9.0\%$, $3.1\%$, and $22.2\%$ over the second-best method. These results validate the superior generality of the proposed \textsc{ConTriNet} in handling various unknown challenging scenarios.

Visualization result comparisons are shown in Figure~\ref{fig:VT-IMAG} for the different attributes in VT-IMAG, which represent challenging scenes encountered in autonomous driving. These qualitative results directly demonstrate the robustness of the proposed \textsc{ConTriNet} against common and unknown scenes. It is noteworthy that our method accurately identifies salient objects with complete structures and clear boundaries, even in the presence of unforeseen distortions, such as out-of-focus ($10^{th}$ row), rainy day ($11^{th}$ row), and foggy day ($12^{th}$ row), while other competitors are either sensitive or fail to locate salient objects in the presence of unknown distortions. In fact, for most of the challenging scenes illustrated in Fig.~\ref{fig:VT-IMAG}, the RGB modality provides limited saliency cues, while the thermal modality is instrumental, particularly in adverse weather conditions and low nighttime illumination. The reliable results predicted by our \textsc{ConTriNet} can be attributed to the adopted powerful ``Divide-and-Conquer'' strategy, which efficiently utilizes modality-specific and modality-complementary cues for RGB-T SOD.

\subsection{Ablation Study}
All ablation study experiments are conducted using \textsc{ConTriNet}$^{\tiny{50}}$. Initially, a UNet-like single-flow model is constructed as a naive baseline, where the features from each layer of the encoder in both modalities are directly added. Table~\ref{tab:cmp_MFM} demonstrates the effectiveness of this strong baseline~(referred to as "No.1"), which serves as a reliable foundation for further performance improvements.

\subsubsection{Effectiveness of MFM}
\begin{figure}[tb!]
	\centering
	\begin{overpic}
        [width=1 \linewidth]{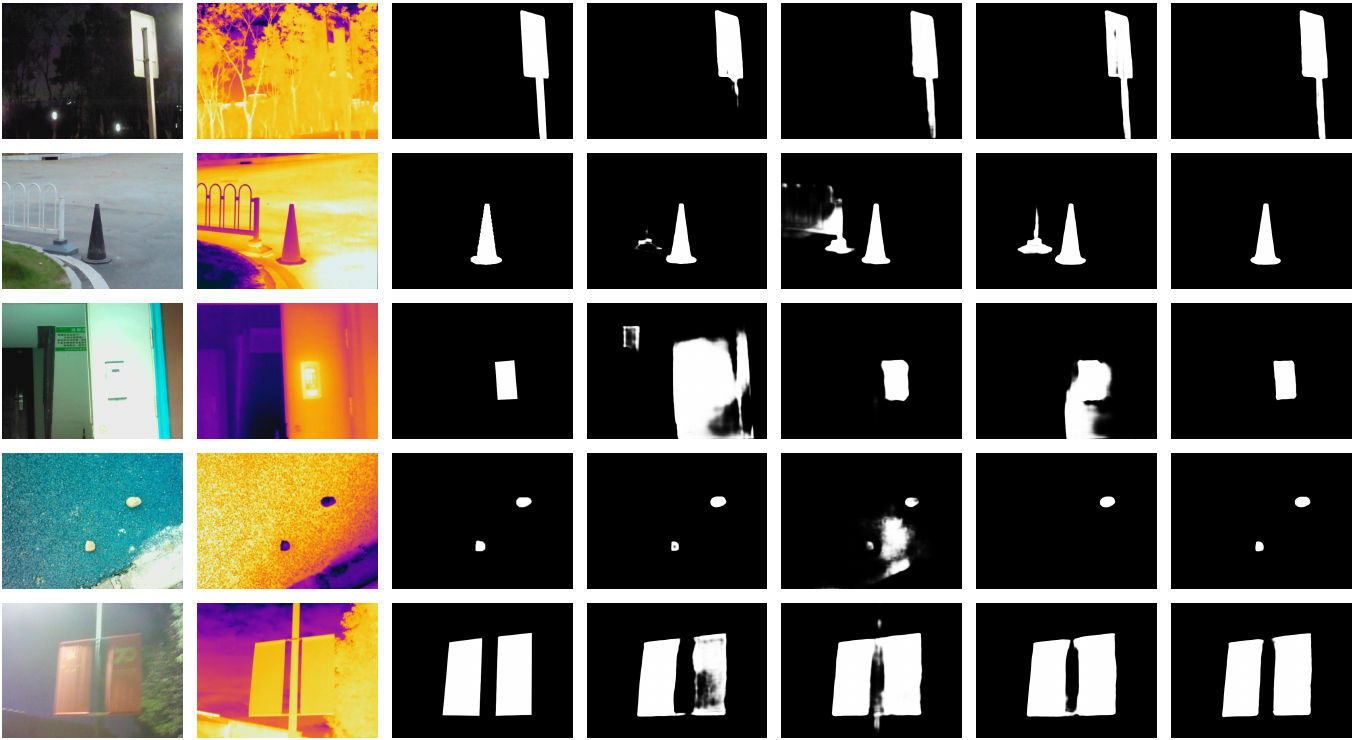}
        \vspace{0.1mm}
        \put(5.5, -3.5){{\scriptsize \textbf{(a)}}}
        \put(19.5, -3.5){{\scriptsize \textbf{(b)}}}
        \put(34.5, -3.5){{\scriptsize \textbf{(c)}}}
        \put(48.5, -3.5){{\scriptsize \textbf{(d)}}}
        \put(62.5, -3.5){{\scriptsize \textbf{(e)}}}
        \put(78.5, -3.5){{\scriptsize \textbf{(f)}}}
        \put(91.5, -3.5){{\scriptsize \textbf{(g)}}}
        \end{overpic}
        \vspace{0.5mm}
	\caption{Visualization examples of ablation studies. (a) RGB map. (b) Thermal map. (c) Ground Truth. (d) \textsc{ConTriNet} (w/o MFM). (e) \textsc{ConTriNet} (w/o RASPM). (f) \textsc{ConTriNet} (w/o MDAM). (g) \textsc{ConTriNet} (full model).}
	\label{fig:abs}
	\vspace{-2mm}
\end{figure}
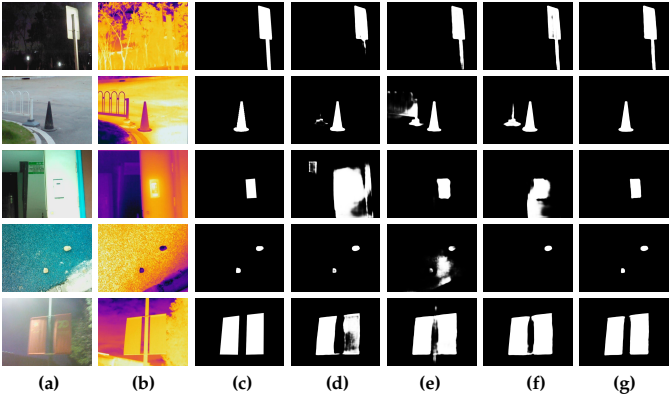

In our proposed \textsc{ConTriNet}, the modality-induced feature modulator (MFM) aims to fully exploit the complementary information between modalities and achieve adaptive fusion of cross-modality features. The MFM process can be divided into two steps: cross-guided feature enhancement (CFE) and attention-aware feature fusion~(AFF). We conduct a thorough ablation analysis to evaluate the interior design choices of MFM, as shown in Table~\ref{tab:cmp_MFM}. Specifically, ``No.2" indicates MFM with only AFF, while ``No.3" indicates MFM with only CFE. The results show that each step individually enhances the SOD performance to some extent.
Comparing ``No.2/3" with ``No.4", we observe that both independent components can synergistically enhance the network's robustness. This could be attributed to the former's ability to exploit the complementary information from different modalities and the latter's capability to adaptively fuse cross-modality features. The consistent performance improvement across all datasets confirms the effectiveness of our structural design. Further evidence of the overall effectiveness of MFM in our \textsc{ConTriNet} is shown in Figure~\ref{fig:abs} (\emph{i.e.,~}(d)~\emph{vs.}~(g)). Additionally, in Figure~\ref{fig:vis_feature}, we present an example of feature evolution outputted from MFM, demonstrating that the proposed modality-induced feature modulator not only fully exploits the complementary information but also reduces the inherent discrepancy between modalities.

\subsubsection{Superiority of RASPM}
The residual atrous spatial pyramid module (RASPM) serves as the key element of the customized decoders in triple flows. As shown in Table~\ref{tab:cmp_MFM}, compared to ``No.6", the inferior results of ``No.5" underscore the indispensability of RASPM. 
By introducing RASPM, as illustrated in Fig.~\ref{fig:abs} (\ie,~(e)~\emph{vs.}~(g)), salient objects can be highlighted with complete structures and clear boundaries. To further verify the benefits of RASPM, we substitute it with alternative methods for multi-scale feature learning, each having comparable model parameters and computational costs, as illustrated in Table~\ref{fig:RASPB}, specifically, vanilla convolution, PPM~\cite{PPM_17}, and ASPP~\cite{ASPP_18}. 
Notably, ``No.1" denotes a $3\times 3$ convolutional layer incorporating a skip connection similar to ResBlock~\cite{HeZRS16}. PPM~\cite{PPM_17} and ASPP~\cite{ASPP_18},extensively employed in SOD tasks,  exploit separate branches to extend the receptive field and enhance the multi-scale feature representations. 
Despite playing a role akin to RASPM, these modules fall short when integrated into our \textsc{ConTriNet}, yielding suboptimal performance, which strongly affirms the superiority of RASPM. Furthermore, we employ vanilla convolution operations with the identical kernel size in an experiment, referred to as ``No.4", to validate the efficacy of atrous convolution operations. The comparison between ``No.5" and ``No.4" corroborates that our model with the comprehensive version of RASPM achieves superior performance. This phenomenon can be attributed to the ability of atrous convolution to obtain a larger receptive field than ordinary convolution, thus facilitating the extraction of multi-scale contextual information.

\begin{table}[t!]
    \centering
    \scriptsize
    \caption{\label{tab:decoder}Performance comparisons of saliency maps predicted by different flows on three datasets.}
    \vspace{-1mm}
    \label{tab:cmp_salmaps}%
    \setlength\tabcolsep{0.3em}
    \resizebox{\linewidth}{!}{%
    \begin{tabular}{l*{3}{|*{4}{c}}}
	\toprule[2pt]
    & \multicolumn{4}{c|}{\textbf{VT821}} & \multicolumn{4}{c|}{\textbf{VT1000}} & \multicolumn{4}{c}{\textbf{VT5000}}                           \\
    \multirow{-2}{*}{\textbf{Metric}}             
	& $M_{r}$ 
	& $M_{t}$
	& $M_{s}$
	& $M_{f}$
	& $M_{r}$ 
	& $M_{t}$
	& $M_{s}$
	& $M_{f}$
	& $M_{r}$ 
	& $M_{t}$
	& $M_{s}$
	& $M_{f}$
	\\ 
	\midrule[1pt]
	$\mathcal{M}\downarrow$ & 0.045 & 0.058 & 0.031 & \textbf{0.027} & 0.026 & 0.047 & 0.023 & \textbf{0.019} & 0.038 & 0.059 & 0.033 & \textbf{0.030}
	\\
	$E_{m}\uparrow$ & 0.887 & 0.877 & 0.920 & \textbf{0.926} & 0.932 & 0.905 & 0.940 & \textbf{0.948} &0.916 & 0.878 & 0.929 & \textbf{0.934}
	\\
	$F_{\beta}\uparrow$ & 0.781 & 0.742 & 0.833 & \textbf{0.850} & 0.873 & 0.820 & 0.885 & \textbf{0.906} & 0.826 & 0.742 & 0.848 & \textbf{0.863}
	\\
	$S_m\uparrow$ & 0.861 & 0.816 & \textbf{0.894} & 0.892 & 0.922 & 0.872 & \textbf{0.929} & \textbf{0.929} & 0.882 & 0.818 & 0.893 & \textbf{0.894}                              
	\\
	$F^{\omega}_{\beta}\uparrow$ & 0.751 & 0.691 & 0.812 & \textbf{0.840} & 0.870 & 0.794 & 0.880 & \textbf{0.907} & 0.806 &0.697&0.816&\textbf{0.848}
	\\
	\bottomrule[2pt]
\end{tabular}

      }
    \vspace{-2mm}
\end{table}%

\subsubsection{Effectiveness of MDAM}
The modality-aware dynamic aggregation module (MDAM) is proposed to integrate informative cues excavated from RGB- and Thermal-modality flows into the modality-complementary flow.
To evaluate its impact, we conduct an ablation study by removing MDAM from the modality-complementary flow, referred to as ``No.6" in Table~\ref{tab:cmp_MFM}. The results, compared to ``No.9", demonstrate the significant role of MDAM in fusing complementary features across different modalities at multiple layers, leading to improved saliency results. 
Furthermore, to validate the efficacy of the dynamic aggregation mechanism, we set the learnable dynamic weights, $\alpha$ and $\beta$, to a fixed value of $1$ in ``No.7". While this fixed strategy outperforms others on VT1000, it fares worse than the dynamic strategy on the more challenging VT821 and VT5000. This discrepancy arises due to the failure of the fixed fusion strategy to consider the contributions of different modalities when handling low-quality data. 
To further elucidate the impact of each component within MDAM, we introduce a new baseline model ``No.8'', which excludes the Detail-oriented Enhancement (DoE) while retaining other dynamic elements of MDAM. The results show that excluding DoE slightly decreased the effectiveness of MDAM across all evaluated metrics, confirming the additional value of DoE in refining RGB features to enhance the saliency detection capability of our model.
These findings emphasize the importance of dynamically aggregating modality-aware features for accurate salient object localization. Additionally, to provide a more intuitive understanding of MDAM's overall effect in our \textsc{ConTriNet}, we visualize the results of the ablation study in Figure~\ref{fig:abs} (\emph{i.e.,~}(f)~\emph{vs.}~(g)).

\subsubsection{Discussion of Loss Functions}
In our model, we employ the hybrid loss which consists of the weighted binary cross entropy (wBCE) loss $\mathcal{L}_{bce}^{w}$ and the weighted IoU (wIoU) loss $\mathcal{L}_{iou}^{w}$ as mentioned in Section~\ref{sec:loss_function}. To assess its efficacy, we evaluate the performance of training with a single loss function, specifically the wBCE loss alone or the wIoU loss alone. Table~\ref{tab:cmp_loss} demonstrates that the hybrid loss is a crucial factor in the success of our \textsc{ConTriNet} for achieving robust performance. The wBCE loss helps to improve $S_m$ and $\mathcal{M}$, while the wIoU loss contributes to promoting the score of $F_{\beta}$ and $E_m$. All things considered, we adopt the hybrid of the wBCE loss and wIoU loss as our default setting to optimize the proposed \textsc{ConTriNet}.

\begin{figure}[t!]
\centering
\begin{tabular}{c}
\includegraphics[width=0.99\linewidth]{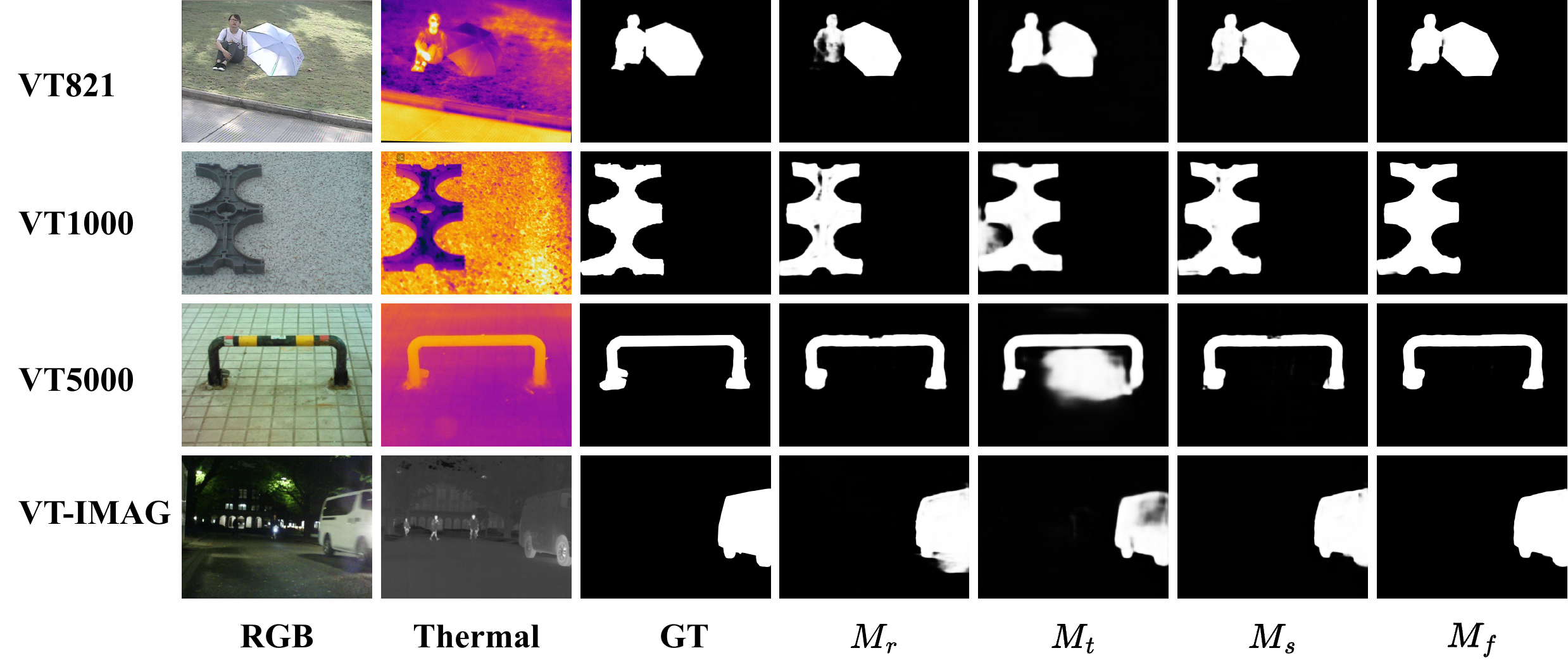}
\end{tabular}
\vspace{-2mm}
\caption{Comparative visualizations of saliency maps predicted by the different flows of the model (zoomed-in for a better comparison).}%
\label{fig:decoders}
\vspace{-2mm}
\end{figure}  

\begin{table*}[t!]
    \centering
    \caption{Comparison of the proposed \textsc{ConTriNet} with different framework configurations on three datasets.}
    \vspace{-1mm}
    \label{tab:cmp_framework}%
    \resizebox{\linewidth}{!}{%
    \begin{tabular}{c|l*{3}{|*{5}{c}}|c}
	\toprule[2pt]
    & & \multicolumn{5}{c|}{\textbf{VT821}} & \multicolumn{5}{c|}{\textbf{VT1000}} & \multicolumn{5}{c|}{\textbf{VT5000}}                                                                                  \\
	\multirow{-2}{*}{\textbf{No.}}
	&\multirow{-2}{*}{\textbf{Settings}}            
	& $S_{m}~\uparrow$ 
	& $F_{\beta}~\uparrow$  
	& $F^{\omega}_{\beta}~\uparrow$   
	& $E_{m}~\uparrow$ 
	& $\mathcal{M}~\downarrow$                             
	& $S_{m}~\uparrow$ 
	& $F_{\beta}~\uparrow$  
	& $F^{\omega}_{\beta}~\uparrow$   
	& $E_{m}~\uparrow$ 
	& $\mathcal{M}~\downarrow$                     
	& $S_{m}~\uparrow$ 
	& $F_{\beta}~\uparrow$  
	& $F^{\omega}_{\beta}~\uparrow$   
	& $E_{m}~\uparrow$ 
	& $\mathcal{M}~\downarrow$   
        % & \multirow{-2}{*}{\textbf{FLOPs (G)}} 
        & \multirow{-2}{*}{\textbf{Params. (M)}}
	\\ 
	\midrule[1pt]
	1 & Modality-shared Encoder ($\times$1)                    
	&{\textbf{0.892}}   &{\textbf{0.850}}    &{\textbf{0.840}}    &{\textbf{0.926}}    &{\textbf{0.027}}   
	& 0.929   & 0.906    & 0.907    & 0.948    & 0.019 
	& 0.894   &{\textbf{0.863}}    & 0.848    &{\textbf{0.934}}   & 0.030 
        & 34.77
	\\
	2 & Modality-specific Encoder ($\times$2)                          
	& 0.891 & 0.845 & 0.835 & 0.923 & 0.029
	& {\textbf{0.932}} & {\textbf{0.907}} & {\textbf{0.908}} & {\textbf{0.949}} & {\textbf{0.018}} 
	& {\textbf{0.897}} & 0.862 & {\textbf{0.852}} & 0.933 & {\textbf{0.029}} 
        & 58.44
	\\
	\midrule
        \midrule
	3 & Modality-complementary Flow ($\times$1) 
	& 0.871 & 0.812 & 0.798 & 0.900 & 0.037
	& 0.928 & 0.893 & 0.900 & 0.943 & 0.021
	& 0.887 & 0.836 & 0.825 & 0.920 & 0.034
        & 27.67
	\\
	4 & Modality-specific Flow ($\times$2) 
	& 0.883 & 0.835 & 0.820 & 0.912 & 0.033
	& 0.927 & 0.900 & 0.898 & 0.944 & 0.022
	& 0.889 & 0.852 & 0.833 & 0.929 & 0.032
        & 30.16
	\\
	5 & Full Flow Integration ($\times$3) 
	&{\textbf{0.892}}   &{\textbf{0.850}}    &{\textbf{0.840}}    &{\textbf{0.926}}    &{\textbf{0.027}}   
	&{\textbf{0.929}}   &{\textbf{0.906}}    &{\textbf{0.907}}    &{\textbf{0.948}}    &{\textbf{0.019}} 
	&{\textbf{0.894}}   &{\textbf{0.863}}    &{\textbf{0.848}}    &{\textbf{0.934}}   &{\textbf{0.030}}
        & 34.77
	\\
	\bottomrule[2pt]
\end{tabular}

      }\vspace{-2mm}
\end{table*}%

\subsubsection{Analysis of Flow-Cooperative Fusion}
As mentioned in Sec.~\ref{sec:loss_function}, the triple flows in \textsc{ConTriNet} generate separate saliency maps denoted as $M_{r}, M_{t}$, and $M_{s}$ using the ``Divide-and-Conquer'' strategy. 
To improve the accuracy and resilience of \textsc{ConTriNet} in complex scenarios, a \textit{flow-cooperative fusion strategy} is employed to obtain a high-quality saliency map for the final prediction.
Table~\ref{tab:decoder} presents quantitative comparisons of various saliency maps and their combinations. Although $M_{r}, M_{t}$, and $M_{s}$ have their own advantages and disadvantages in different regions, our final $M_{f}$ effectively combines their advantages and suppresses the disadvantages, resulting in improved results with sharper edges and complete structure. This highlights the advantageous nature of the proposed flow-cooperative fusion strategy for achieving accurate segmentation of common salient objects. 
Moreover, the qualitative comparisons in Figure~\ref{fig:decoders} demonstrate that the final RGB-T saliency performance is significantly improved by the effective flow-cooperative fusion strategy, which suppresses irrelevant regions and improves indistinct predictions.

\subsubsection{Evaluation of Framework Configurations}
As shown in Table~\ref{tab:cmp_framework}, we conduct an extensive ablation study to assess the impacts of various framework configurations within our \textsc{ConTriNet}. Specifically, we analyze a modality-shared encoder that simplifies the model by using a unified encoder for all modalities (``No.1''), contrasting it with a dual-encoder setup for separate modality processing (``No.2''). Additionally, we investigate various decoder configurations: a single modality-complementary flow (``No.3''), dual modality-specific flows (``No.4''), and the full integration of all flows (``No.5''). Comparing ``No.1'' and ``No.2'' reveals that the modality-shared encoder not only achieves comparable or superior performance to the dual encoders, especially demonstrating notable superiority on the VT821 dataset, but also significantly reduces the number of model parameters (\emph{i.e.,}~34.77M \emph{vs.} 58.44M). Regarding decoder configurations, while ``No.3'' provides a solid baseline, integrating modality-specific flows in ``No.4'' substantially enhances model performance, particularly in precision and error metrics. The comprehensive integration of all flows in ``No.5'' consistently yields the highest performance across all evaluated metrics and datasets. This confirms the effectiveness of our ``Divide-and-Conquer'' strategy in improving the overall robustness and accuracy of the framework, thereby demonstrating its potential to adeptly handle complex RGB-T SOD tasks.

\subsection{Failure Cases}
In the preceding sections, the efficacy and robustness of the proposed \textsc{ConTriNet} have been demonstrated via diverse quantitative and qualitative experiments. While the \textsc{ConTriNet} achieves satisfactory results for robust RGB-T SOD, further improvements can still be made. Fig.~\ref{fig:failure_case} illustrates several typical failure cases. Notably, in highly complex scenes with abundant interference information, \textsc{ConTriNet} fails to accurately locate salient objects. For example, in case (a), where there is overexposure, the model struggles to extract salient objects due to the intense lights emitted by an oncoming car in the dark.
Additionally, in scenarios with intricate backgrounds (\eg~case (b) and (c)), our model also falls short in capturing salient objects with complete structure. In case (b), our triple-flow network with the ``Divide-and-Conquer'' strategy successfully localizes the objects but is unable to produce flawless segmentation, which necessitates improved discrimination of finer object structures and boundary details. Case (d) reveals that our model tends to predict relatively large salient objects and heavily relies on the quality of the input RGB-T image pair, resulting in erroneous salient region predictions when thermal information is misleading. In conclusion, the challenges of \textit{uncertainty}, \textit{ambiguity}, and \textit{misleadingness} in real-world robust RGB-T SOD merit further research.

\newcommand{\AddImge}[1]{%
    \includegraphics[width=.1\textwidth]{Imgs/Failure/#1_RGB.png} &%
    \includegraphics[width=.1\textwidth]{Imgs/Failure/#1_T.png} &%
    \includegraphics[width=.1\textwidth]{Imgs/Failure/#1_GT.png} &%
    \includegraphics[width=.1\textwidth]{Imgs/Failure/#1_Ours.png} %
}

\begin{figure}[t!]
    \centering
    \footnotesize
    \resizebox{0.45\textwidth}{!} {
    \begin{tabular}{m{0.1cm}<{\centering}m{1.5cm}<{\centering}m{1.5cm}<{\centering}m{1.5cm}<{\centering}m{1.5cm}<{\centering}}
        (a)&\AddImge{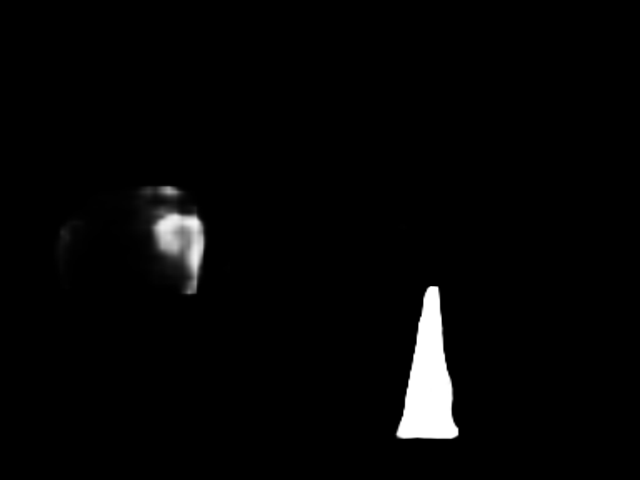} \\
        (b)&\AddImge{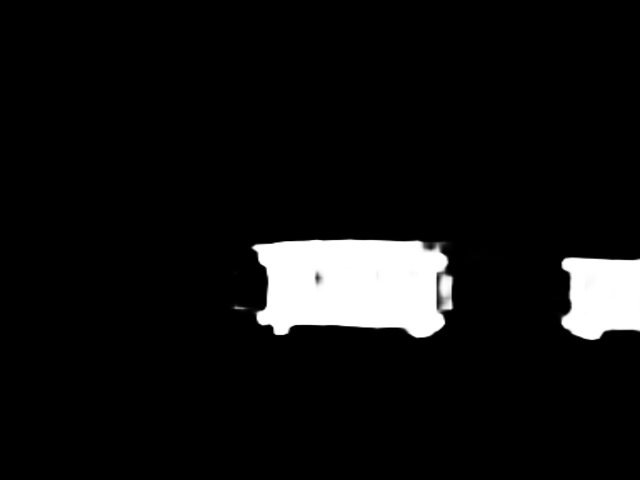} \\
        (c)&\AddImge{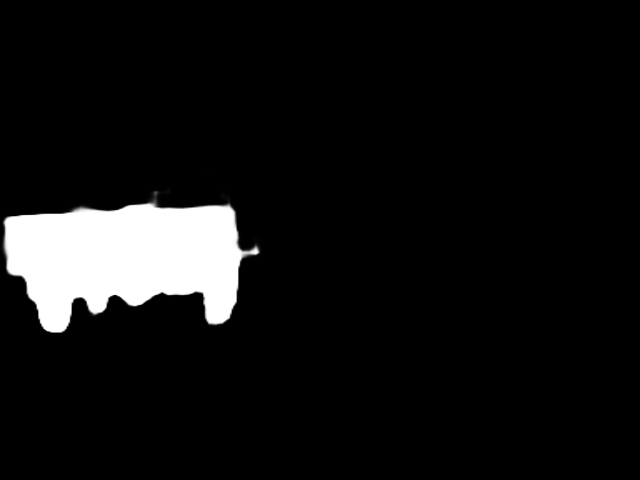} \\
        (d)&\AddImge{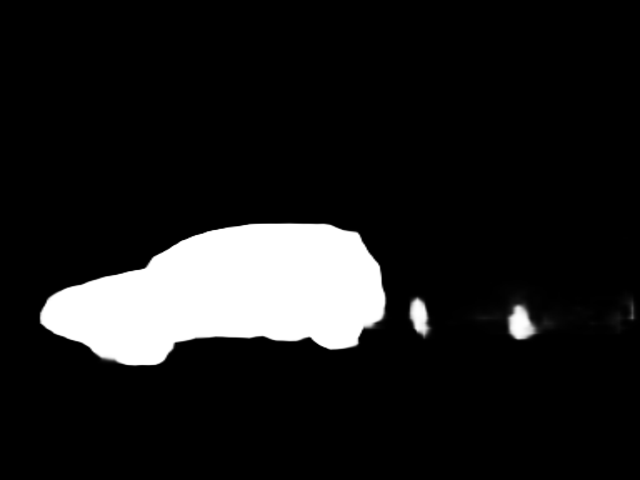} \\
        & {\textbf{RGB}} & {\textbf{Thermal}} & {\textbf{GT}} & {\textbf{Ours}}
    \end{tabular}
    }
    \caption{Failure cases from the proposed VT-IMAG (zoomed-in for a better comparison).}
    \vspace{-2mm}
    \label{fig:failure_case}
\end{figure}

\section{Conclusion}
\label{section5}
In this paper, we propose \textsc{ConTriNet}, a novel confluent triple-flow network that incorporates the ``Divide-and-Conquer'' strategy into the robust framework for the RGB-T SOD task. \textsc{ConTriNet} consists of a unified encoder alongside three specialized decoders that address different subtasks, including learning the saliency-oriented modality-specific representation and modality-complementary representation, thus enabling a more comprehensive and resilient perception for RGB-T SOD. The well-designed components in \textsc{ConTriNet} exhibit a strong ability to dynamically integrate modality-complementary information and also to focus on the deep excavation of modality-specific information, reducing the interference of defective modality, which results in high robustness to challenging scenes. Furthermore, to evaluate the robustness and generality of \textsc{ConTriNet}, we construct a new comprehensive RGB-T SOD benchmark dataset named VT-IMAG, with various challenging scenarios, to serve as a testbed for verifying the robustness of different models. Extensive experimental results demonstrate the natural strengths of the proposed \textsc{ConTriNet} in handling SOD problems in extremely challenging scenarios. In the future, it would be appealing to extend \textsc{ConTriNet} to a flexible architecture that can accommodate RGB-X data and enhance SOD performance in challenging scenarios.

% --------------------------------------------

% --------------------------------------------
{
\bibliographystyle{IEEEtran}
\bibliography{TPAMI_SOD}
}

% --------------------------------------------
\end{document}